%% file: main.tex
\documentclass[11pt]{article}

\usepackage[preprint]{acl}
\ifdefined\XeTeXversion
  \usepackage{fontspec}
\else
  \usepackage{times}
  \usepackage[T1]{fontenc}
  \usepackage[utf8]{inputenc}
  \usepackage{inconsolata}
\fi
\usepackage{latexsym}
\usepackage{microtype}
\usepackage{graphicx}
\usepackage{booktabs}
\usepackage{amsmath}
\usepackage{amssymb}
\usepackage{mathtools}
\usepackage{amsthm}
\usepackage[capitalize,noabbrev]{cleveref}
\usepackage{xcolor}
\usepackage{multirow}

\newcommand{\cond}[1]{\textsf{#1}}
\newcommand{\pp}{\,pp}
\newcommand{\figorbox}[2]{\includegraphics[width=#2]{#1}}
\title{Context Has a Price: Adding and Removing Context\\ for LLM Coding Agents under Prefix Caching}
\input{authors}
\date{}
\hypersetup{pdfauthor={Zerui Yang, Miaomiao Dong, Quan Xue, Chaorui Zhang, Xuanyu Zheng, Xiyuan Wang, Haolin Ruan, Xueyan Niu, Guangcheng Song, Weixi Zhang, Wei Han, Bo Bai},pdftitle={Context Has a Price: Adding and Removing Context  for LLM Coding Agents under Prefix Caching}}
\begin{document}
\maketitle
\input{sections/00_abstract}
\input{sections/01_intro}
\input{sections/02_pricing}
\input{sections/03_setup}
\input{sections/04_adding}
\input{sections/05_removing}
\input{sections/06_reeval}
\input{sections/07_related}
\input{sections/08_discussion}
\input{sections/09_limitations}
\input{sections/10_impact}

\bibliography{references}

\appendix
\input{sections/appendix}

\end{document}

%% file: authors.tex
\author{
  \textbf{Zerui Yang}\thanks{Corresponding author}, \textbf{Miaomiao Dong}, \textbf{Quan Xue}, \textbf{Chaorui Zhang}\thanks{Corresponding author}, \\
  \textbf{Xuanyu Zheng}, \textbf{Xiyuan Wang}, \textbf{Haolin Ruan}, \textbf{Xueyan Niu}, \\
  \textbf{Guangcheng Song}, \textbf{Weixi Zhang}, \textbf{Wei Han}, \textbf{Bo Bai} \\
  {\normalfont Huawei Technologies Ltd.} \\
  {\normalfont\small Correspondence:
  \texttt{richard.yangpolyu@gmail.com},
  \texttt{zhang.chaorui@huawei.com}}
}

%% file: sections/00_abstract.tex
\begin{abstract}
Coding agents repeatedly send their conversation history to a language model. Adding useful code may reduce search, while removing old tool outputs may shorten later prompts. But do these changes reduce the cost of solving a task? We study this as a measurement problem in more than 20{,}000 runs of open models on SWE-bench Verified, SWE-bench and Loc-Bench. We account separately for uncached input, cached input and output tokens. Three findings emerge. First, in our main evidence-package experiments, retrieved code reduces read and search commands by 18--34\%, but increases cost by 33--42\% when cached input costs one tenth of uncached input, without a measurable improvement in resolution. Second, removing old observations can either save money or increase cost: the same approach saves 19--36\% without a cache discount but increases cost by 38--116\% with a tenth-price cache. Third, shorter prompts do not guarantee cheaper runs because agents may respond by taking more steps. We also find that concurrent experiments can share cached prefixes and distort cost comparisons. These results show why context-management methods should be evaluated using both token prices and the agent's subsequent behavior.
\end{abstract}

%% file: sections/01_intro.tex
\section{Introduction}
\label{sec:intro}

\begin{figure*}[t]
\centering
\includegraphics[width=\textwidth]{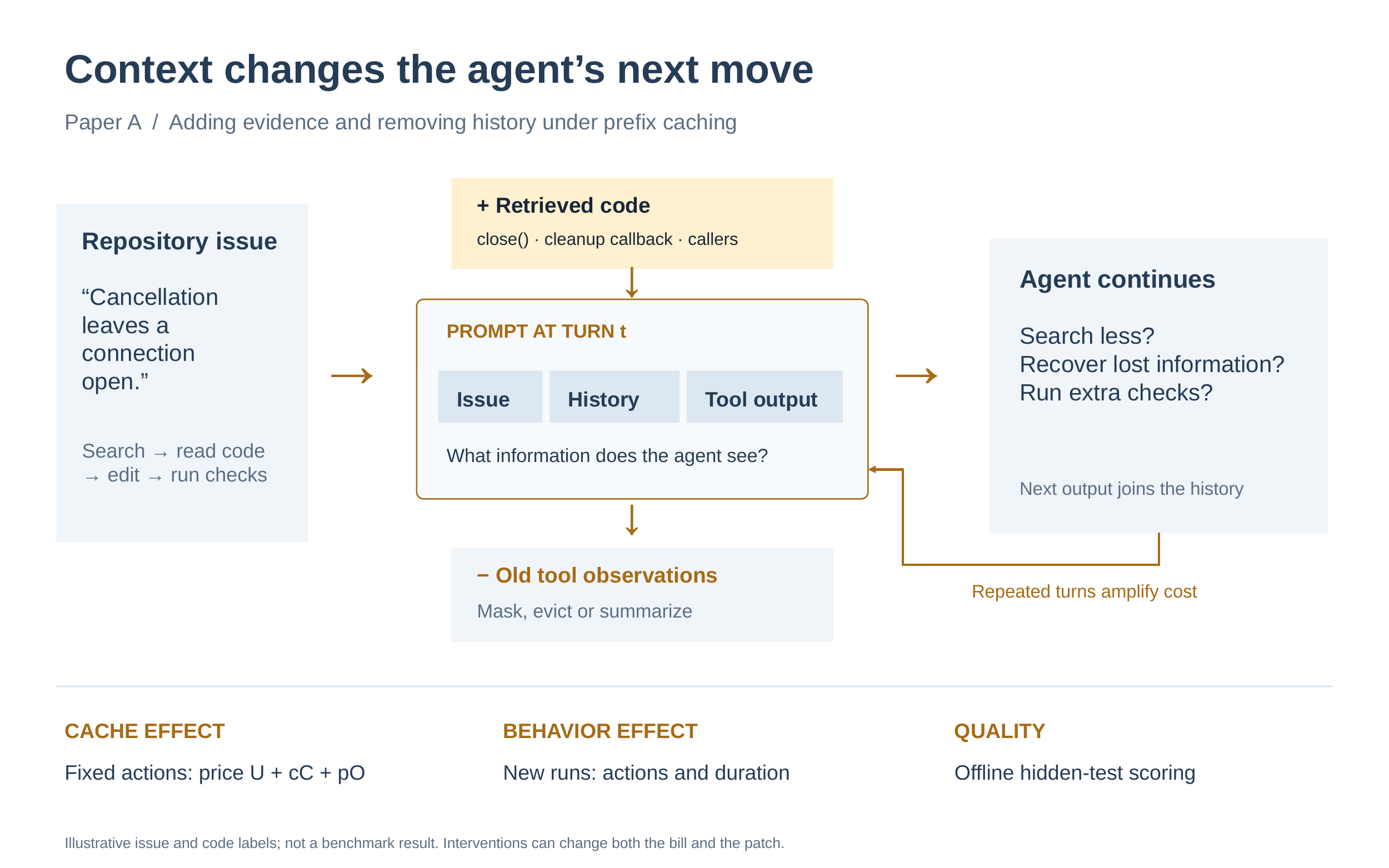}
\caption{Adding and removing context can change both the bill and the patch. Retrieved code adds evidence to the prompt; masking, eviction or summarization removes old observations. Both interventions can affect the agent's later actions. We separate direct cache effects under fixed actions from behavioral responses in new runs, and evaluate patch quality offline. The issue and code labels are illustrative.}
\label{fig:overview}
\end{figure*}

Coding agents solve repository-level tasks by searching for relevant code, editing files and running checks \citep{jimenez2024swe,yang2024swe,wang2025openhands}. Each action adds information to the conversation. As a run continues, the model repeatedly reads this growing history. Context management therefore presents a practical question: should we give the agent more information to guide its work, or remove old information to reduce its cost? Figure~\ref{fig:overview} summarizes the setting.

Both choices appear useful. Retrieved code may save the agent several searches. Removing old tool outputs may make later prompts shorter. However, the number of tokens alone does not determine the bill. Serving systems reuse cached prompt prefixes \citep{kwon2023efficient,zheng2024sglang}, and cached input is often priced below fresh input. Output tokens have a separate price. We therefore calculate cost as
\begin{equation}
\mathrm{cost}(c, p) = U + c\,C + p\,O,
\label{eq:cost}
\end{equation}
where $U$ and $C$ are the uncached and cached input tokens summed over all requests, $O$ the output tokens, and $c$ and $p$ the prices of a cached input token and an output token relative to an uncached input token.

Two examples explain why these prices matter. A token added to the prompt is paid again at every later turn. Changing an earlier part of the prompt invalidates the cache from that point on, so the rest of the prompt is paid at the uncached price once more. Methods that reduce token counts can thus raise cost \citep{weinberger2026token}, and whether they do depends on prices that papers seldom report. We ask whether adding or removing context reduces the total cost of solving a task. We treat adding and removing context as two tests of the same accounting question: whether a local change to the prompt lowers total inference cost once cache reuse and subsequent actions are included. We compare both approaches in controlled experiments with prespecified tests (\cref{tab:summary}):
\begin{itemize}
\item \textbf{Adding context.} Retrieved code context, such as candidate edit locations and related definitions, is meant to make agents search less, spend less and fix more \citep{chen2025locagent,ouyang2025repograph,xia2024agentless}.
\item \textbf{Removing context.} Masking or evicting old observations shortens the prompt \citep{lindenbauer2025complexity,xiao2026reducing,xu2026tokenpilot}.
\end{itemize}
We measure token-level prefix overlap with the model's tokenizer (\cref{sec:pricing}). The experiments use a minimal bash agent and mini-SWE-agent \citep{yang2025minisweagent}, sandboxes with and without the ability to run the repository's tests, and seven open models from three families served locally, over more than 20{,}000 runs (\cref{sec:setup}).

\begin{table*}[t]
\caption{Adding and removing context at a glance. Entries are paired point estimates over tasks. Quality entries are percentage-point changes in resolution unless marked Acc@10; localization accuracy and resolution are different outcomes and should not be compared directly. Cost follows \cref{eq:cost} with $p=4$, counting only cache hits a run could have produced itself (\cref{sec:pricing}). ``Official'' is the SWE-bench harness; ``local'' runs the same hidden tests in our executable sandbox.}
\label{tab:summary}
\centering
\small
\resizebox{\textwidth}{!}{%
\begin{tabular}{@{}llrrrl@{}}
\toprule
Lever & Comparison (setting) & Read+search & Cost, $c=0.1$ & Cost, $c=1$ & Quality \\
\midrule
Add & evidence package vs.\ none (Verified, 80 tasks) & $-32\%$ & $+36\%$ & $+25\%$ & resolved $+3.8$ official \\
Add & evidence package vs.\ none, agent can run tests (183 tasks) & $-9\%$ & $+21\%$ & -- & resolved $-0.3$ official \\
Add & retrieval appended to the agent's localization (184 tasks, 3 models) & -- & $0\%$ & $0\%$ & function Acc@10 $+14$ to $+20$ \\
Remove & rolling masking vs.\ none (201 fresh tasks; 27B / Coder) & -- & $+116\%$ / $+87\%$ & $-19\%$ / $-25\%$ & $-0.5$ / $+2.0$ official \\
Remove & break-even masking ($c=0.1$) vs.\ none (same) & -- & $+2.8\%$ / $+1.8\%$ & $-26\%$ / $-21\%$ & equivalent, $+1.3$ official \\
Remove & rolling / break-even masking vs.\ none, SWE-agent (115 tasks; 27B) & -- & $+101\%$ / $+9.6\%$ & $-29\%$ / $-25\%$ & $-0.9$ / $-1.7$ local \\
Remove & rolling / break-even masking vs.\ none, GLM-4.7-Flash (115 tasks) & -- & $+38\%$ / $-13\%$ & $-36\%$ / $-38\%$ & $+3.5$ / $+7.8$ local \\
Remove & rolling / break-even masking vs.\ none, Gemma-4 (same) & -- & $+56\%$ / $+14\%$ & $-19\%$ / $-20\%$ & $+6.2$ / $+8.8$ local \\
Remove & LLM summarization vs.\ none (same; 27B / Coder / Qwen3.8) & -- & $+22\%$ / $+28\%$ / $-19\%$ & $-14\%$ / $+3\%$ / $-53\%$ & $+4.2$ / $-0.5$ / $-7.8$ official \\
Remove & reflection (AgentDiet-style) vs.\ none (same) & -- & $+36\%$ / $+31\%$ / $+29\%$ & $+4\%$ / $+7\%$ / $+2\%$ & $-4.2$ / $+5.2$ / $-4.1$ official \\
\bottomrule
\end{tabular}}
\end{table*}

\paragraph{Findings.}
\begin{itemize}
\item \textbf{The tested evidence package reduces search without an established resolution gain.} A retrieved evidence package cuts read and search commands by 18--34\% on every benchmark and Qwen model we test, and random code of the same length does not. Yet runs cost 33--42\% more even when cached tokens cost a tenth, and the resolved rate does not change measurably. No gain is established when the agent can run the repository's tests either. Retrieval helps as a complement instead: appended to the agent's own answer in localization, it raises function Acc@10 by 14--20 points for three models (\cref{sec:adding}).
\item \textbf{Whether removing context pays depends on the price.} Masking all but the latest ten observations, the setting recommended by \citet{lindenbauer2025complexity}, saves 19--25\% without a cache discount but costs 116\% and 87\% more for two models when cached tokens cost a tenth, and the sign flips in SWE-agent and for GLM-4.7-Flash as well. A closed-form break-even rule halves the premium and saves 10--26\% once cached tokens cost a quarter or more. With the agent's actions held fixed, batched and break-even masking would save Qwen3.6-27B 3--22\% at a tenth-price cache, but the agent reacts to missing information by running longer: none of 33 combinations of method, model and scaffold is cheaper than doing nothing across the range of current API prices, although digest masking saves 17\% at $c=0.1$ for GLM-4.7-Flash, whose runs are long (\cref{sec:removing}). The cost ranking of six methods at a tenth-price cache also differs from the ranking without a discount (\cref{sec:reeval}).
\item \textbf{Concurrent experiments can share cached prefixes.} In ablations whose conditions share prompt prefixes and run side by side, server-reported cache hits include prefixes warmed by sibling conditions. On Verified, removing them changes the cost premium of the evidence package from $+8\%$ to $+36\%$ (\cref{sec:pricing}).
\end{itemize}

\paragraph{Contributions.}
\begin{enumerate}
\item A price-vector accounting for agent runs that reports the three token streams separately and counts a run's own cache hits at token granularity. It exposes cache sharing between concurrent conditions as a source of bias.
\item Preregistered evidence on adding and removing context across benchmarks, scaffolds and models, including confirmations on tasks never used before.
\item A fixed-action break-even baseline for removing context, a replay that splits a method's cost into the price of breaking the cache and the agent's reaction, and a re-evaluation of six methods for three models under several price vectors.
\item An evaluation of four further strategies: a unified context controller, a break-even rule that charges for longer runs, truncation at the source and narrow-read training. We also find that appending retrieval results to the agent's final localization improves localization for all three models tested.
\end{enumerate}

%% file: sections/02_pricing.tex
\section{Pricing an Agent Run}
\label{sec:pricing}

\paragraph{Three streams, one price vector.}
For a run with requests $t=1,\dots,T$, let $P_t$ be the prompt tokens of request $t$, $H_t\le P_t$ those served from the prefix cache, and $O_t$ the output tokens. Then $U=\sum_t (P_t-H_t)$, $C=\sum_t H_t$ and $O=\sum_t O_t$ in \cref{eq:cost}. We report the three streams separately and price them at $c\in\{0.1,1\}$, that is, with a 90\% cache discount and with none, and with $p=4$. Other prices are in \cref{app:cost}. At $c=0.1$, output accounts for 19--38\% of the cost of our runs, so ignoring it, or ignoring the discount, changes which intervention looks cheaper.

\paragraph{Reference prices and deployment scope.}
The API price lists cited in this study set the cache-read price at 0.025--0.1 of the uncached input price and output at 5--6 times it \citep{anthropic2026pricing,openai2026pricing}, motivating the low-cache-price settings we evaluate. These are reference prices, not measurements of a provider's billing system. As an external comparison, in 8{,}058 Claude Code and Codex sessions, 95.6\% of input tokens were cache reads \citep{zhu2026tracelab}, against 94--98\% in our runs. Priced with \cref{eq:cost} at $c=0.1$ and $p=5$, those sessions split 61\% cached input, 28\% uncached input and 11\% output, close to the 58.6\%, 30\% and 11\% the trace authors report from actual bills. The cached share of the bill is thus the quantity a price change moves most: it is 18--30\% of our runs' cost at $c=0.025$ and 90--95\% at $c=1$.

\paragraph{Own-cache accounting.}
For an isolated-run comparison, we attribute cache reuse only to prefixes computed by the run's own earlier requests. We therefore set $H_t=\min(P_t,\ \mathrm{LCP}_t)$, where $\mathrm{LCP}_t$ is the longest common token prefix of request $t$ and request $t-1$ of the same run. We obtain it by tokenizing each prompt with the serving model's tokenizer. This is an idealized previous-prompt prefix-overlap measure; it does not model expiration or eviction. vLLM caches in blocks, 784 tokens for Qwen3.6-27B, so it reports up to one block less. This counts reuse of the previous prompt only. Qwen3.6-27B's chat template re-renders the history, and in isolated runs its server never reused more than the previous prompt (0 of 27{,}389 requests); Qwen3-Coder keeps the history verbatim and its server also reused the previous reply, which changes no comparison of \cref{sec:removing} by more than 12 points (\cref{app:cost}). For experiments run before we recorded token ids, we use $\min(\hat H_t, P_{t-1})$, where $\hat H_t$ is the server-reported hit count; where both are available, it counts 9.8\% fewer hits than the token count for Qwen3.6-27B, mostly because of its 784-token blocks, and 0.8\% fewer for Qwen3-Coder.

\paragraph{Concurrent conditions share caches.}
Server-reported hits are not a property of the run alone. In a typical ablation, the conditions of a task share a sampling seed and run side by side on one server. Two conditions whose prompts agree up to some point therefore send identical requests, and the later one reuses the prefix the other has just computed. In our evidence experiments this added 3.7--6.7\% of prompt tokens to the cached share of the affected conditions and none to the baseline. It turned the measured cost premium of the evidence package on Verified from $+8\%$ into $+36\%$ once removed. Run alone on a fresh server, each condition's own counts give $+29\%$ (preregistered). Cost comparisons should use own-cache hits, report the uncached price alongside, and avoid running conditions that share prefixes side by side; \cref{app:protocol} lists the biases we measured and a checklist.

\begin{figure*}[t]
\centering
\figorbox{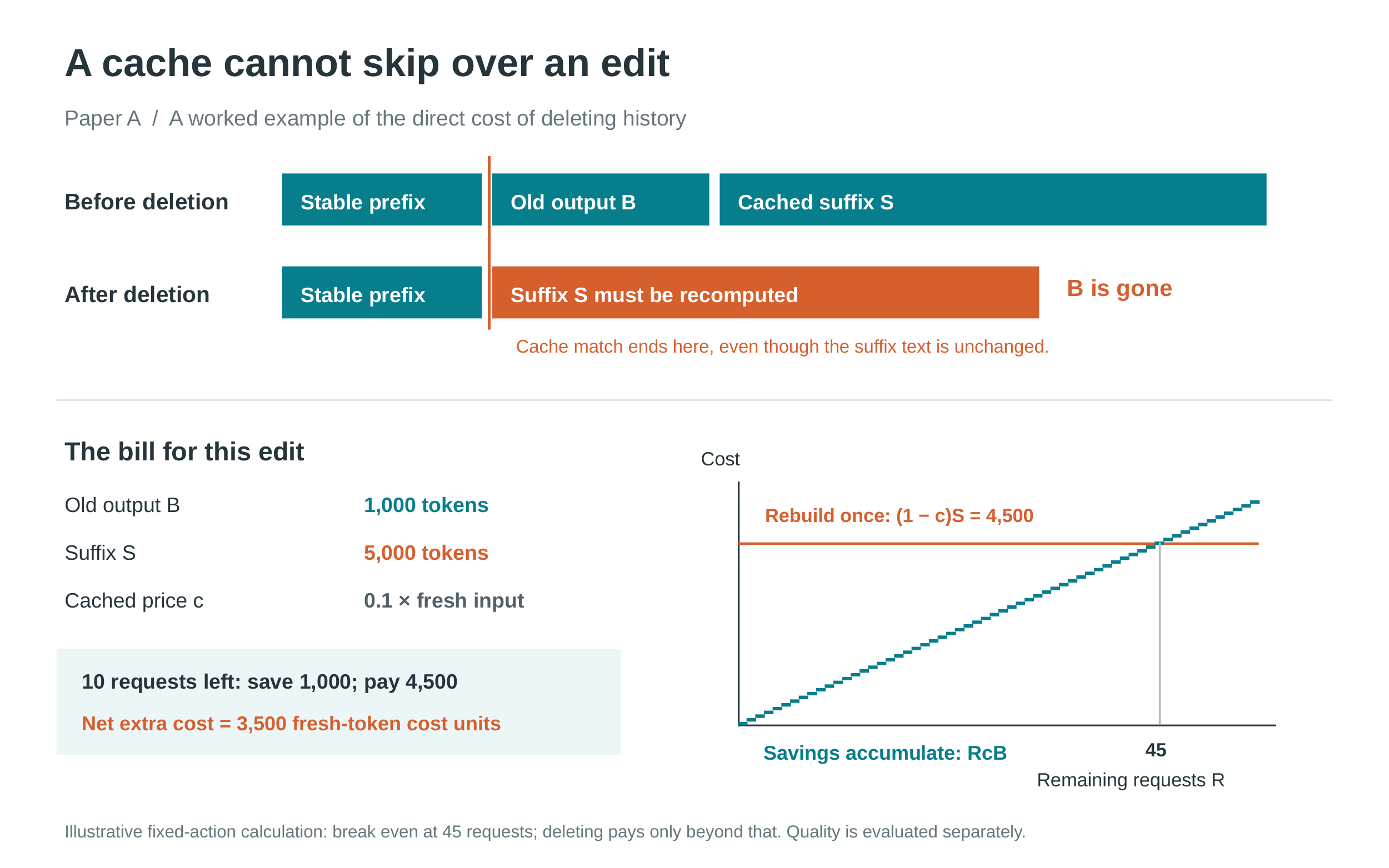}{\textwidth}
\caption{Deleting an old output $B$ shortens later prompts but breaks the cache match for the following suffix $S$. In this illustrative fixed-action example, $B=1{,}000$, $S=5{,}000$ and $c=0.1$. Ten remaining requests save 1{,}000 fresh-token cost units but incur 4{,}500 units to rebuild the suffix; breaking even requires 45 requests. Prompt lengths are schematic, and patch quality is evaluated separately.}
\label{fig:cachebreak}
\end{figure*}

Figure~\ref{fig:cachebreak} gives a worked example of the cache cost.

\paragraph{When does changing the prompt pay?}
Removing $B$ tokens from the prompt before request $t$ saves their cached price $cB$ on each of the $R$ remaining requests. Because the removal changes the prefix, the $S$ tokens of the prompt after the change point lose their cache hit once and cost an extra $(1-c)$ each. Removal pays when
\begin{equation}
R\,c\,B > (1-c)\,S.
\label{eq:breakeven}
\end{equation}
This inequality accounts for token costs with future actions held fixed; it is not a guarantee about an agent that reacts to the edit. The controller can measure $B$ and $S$, but must estimate $R$. At $c=1$ it favors removal when $R>0$ and $B>0$. At $c=0.1$ it favors removing large blocks early and leaving small or late ones. Adding context is the mirror image: $B$ added tokens are paid at least $c\,B$ on every later request. We use \cref{eq:breakeven} both to analyze removal (\cref{sec:adding,sec:removing}) and as a policy: \emph{break-even masking}, which masks all old observations outside the latest ten in one batch whenever the inequality holds.

%% file: sections/03_setup.tex
\section{Experimental Setup}
\label{sec:setup}

\paragraph{Agents and sandboxes.}
We use two main scaffolds so that differences come from the intervention rather than from tooling. Our \emph{minimal agent} gives one bash command per turn in a fenced block, for at most 50 turns or 64K tokens of context. Each command runs in a sandboxed checkout at the task's base commit: separate Linux namespaces, an unprivileged user, no network, per-command time limits, and no installed dependencies. \emph{mini-SWE-agent} 2.4.6 \citep{yang2025minisweagent} keeps its official SWE-bench prompt, loop and submission protocol; we use 250 steps in the executable-sandbox experiments. The earlier evidence-package replication instead used a 50-step cap, as detailed in \cref{app:robustness}. It runs either in the same sandbox, where the repository's tests cannot run, or in an \emph{executable sandbox} with the same isolation plus a read-only Python environment for the task. We built per-version environments for Django~3.0--5.0 and for SymPy following the SWE-bench specifications, without Docker. On 304 Verified tasks, the gold patch passes the hidden tests in 293. The agent cannot install anything. To check the main result in a third scaffold, we also run SWE-agent 1.1.0 \citep{yang2024swe} with its default configuration and function calling; its runtime server runs inside the executable sandbox, and a proxy in front of the model server records every request and tokenizes it exactly as sent. All conditions share the prompt, the sampling settings (temperature 0.7, top-$p$ 0.8, at most 2{,}048 output tokens per turn, thinking disabled) and a seed fixed per task and repeat (\cref{app:agent}).

\paragraph{Benchmarks and models.}
We use 80 SWE-bench test tasks outside Verified \citep{jimenez2024swe}, SWE-bench Verified \citep{openai2024verified}, and 224 Loc-Bench tasks \citep{chen2025locagent}. On Verified, each confirmatory experiment draws tasks never used to derive its hypothesis: 219 fresh tasks for evidence and 201 for masking. The main model is Qwen3.6-27B. We add Qwen3-Coder-30B-A3B, Qwen3.5-4B, Qwen3.5-9B and Qwen3.8-27B, and Gemma-4-26B-A4B and GLM-4.7-Flash from two other families. Models are served with vLLM \citep{kwon2023efficient} with prefix caching, at most eight concurrent runs per server.

\paragraph{Quality.}
For experiments in the dependency-free sandbox we score final patches with the official SWE-bench harness. For experiments in the executable sandbox we also report \emph{local} resolution: the same hidden tests, applied as in SWE-bench, run in our executable sandbox, on the tasks whose gold patch passes there. Official and local results are labeled separately and retain their respective task denominators. The additional official comparisons use only matched tasks with passing gold checks and complete reports (\cref{app:officialreturn}); they are not substituted for local results on a different subset. Both score the recorded tracked-file working-tree patch, which can omit staged changes, rather than the submitted patch file. As a cheaper proxy we report whether a patch edits a function changed by the gold patch (\emph{on target}). Localization accuracy and the on-target proxy do not establish bug resolution. Gold patches and tests are used only for scoring.

\paragraph{Protocol.}
Before each experiment we fixed its task set, conditions, primary hypotheses, test and decision rule in dated internal research records. We use ``preregistered'' to mean specified in those records before the corresponding run or analysis, not independently registered in a public registry. We report every primary test, including unsupported ones (\cref{app:prereg}). The unit of analysis is the task. We average the runs of a task within a condition, pair conditions by task, and compute 95\% confidence intervals from 2{,}000 bootstrap resamples over tasks with a fixed seed. Directional primary tests are one-sided and Holm-corrected within an experiment. Non-inferiority tests compare a one-sided bound with a preregistered margin. Failed runs are kept and rerun once with the same seed.

%% file: sections/04_adding.tex
\section{Adding Context: Less Search, Higher Cost}
\label{sec:adding}

\paragraph{Evidence.}
The evidence package is built without access to the gold patch. A dense code-search model ranks functions by the issue text; the top three non-test functions are shown in full and seven more are listed by location. A code-graph retriever adds related code around the candidates, up to 1{,}000 lines (about 14K tokens). The package (\cond{E5}) is replaced by an index of its locations at the first edit or at turn 8. We compare it with no evidence (\cond{E0}), the three candidate functions alone (\cond{EL}, about 1.2K tokens), the package kept to the end (\cond{E2}), and random code of the same length (\cond{E5r}) (\cref{app:package}).

\paragraph{Exploration falls.}
With our agent, the package cuts read and search commands by 18--34\% on every benchmark and Qwen model. On Verified the cut is 6.64 per run ($-32\%$), and the preregistered confirmation on 219 fresh tasks gives $-4.87$ ($-26\%$). Random code of the same length leaves exploration unchanged ($+0.68$), so the effect comes from relevance, not length. The three candidate functions alone deliver a third to a half of the effect. The effect is smaller in mini-SWE-agent ($-6\%$), and it does not replicate over whole runs with Gemma-4-26B-A4B, although both show a saving of about one command over the first eight. In Gemma, writing a scratch script triggers replacement of the package at a median of step 2, so model behavior and evidence exposure are coupled (\cref{app:robustness}).

\paragraph{Cost rises.}
Evidence in the prompt is paid on every request until it is removed. Runs with the package take fewer turns but cost at least 33--42\% more at $c=0.1$ and 25--33\% more at $c=1$ (\cref{tab:app-main}). Keeping it to the end costs $+80\%$ to $+127\%$. By \cref{eq:breakeven}, replacing the full package with its index pays in 91--97\% of runs at $c=0.1$ and saves a quarter of the cost. Replacing the three candidate functions alone pays in only 42--44\% and saves nothing. In these runs, replacing large evidence after use is cheaper than retaining it; small evidence can stay, and it costs within $-12\%$ to $+5\%$ of no evidence.

\paragraph{No resolution gain is established.}
Under the official harness, the package changes the resolved rate by $-1.9$ points on SWE-bench test and $+3.8$ on Verified (13 tasks better, 10 worse). On 217 fresh Verified tasks (preregistered) it is equivalent to no change within $\pm5$ points ($+0.0$). Pooled over four Qwen models it changes the rate by $+2.9$. Oracle edit locations give a positive estimate ($+5.8$), but the sign test does not establish a gain ($p=0.065$). These results concern this package and its exposure schedule; they do not rule out smaller benefits or gains from other retrieval designs.

\paragraph{Execution and localization controls.}
Evidence might pay off only when the agent can check its hypotheses against execution \citep{li2026codetracer,zhao2026failure}. We crossed evidence with the executable sandbox on 183 Django and SymPy tasks in mini-SWE-agent, two runs per cell (preregistered). Under the official harness, evidence with execution changes resolution by $-0.3$ points, and the interaction is $-1.6$, which rules out a positive interaction larger than about three points. None of the three preregistered tests is supported (\cref{app:robustness}), while the package still cuts exploration and costs about 20\% more. Retrieval and the agent are complementary instead. The agent names one or two locations, mostly correctly, while retrieval ranks many. Appending the retrieval ranking to the agent's answer costs no tokens. On 184 held-out Loc-Bench tasks (preregistered) it raised function Acc@10 from 0.386 to 0.538, which is 15.2 points above the agent alone and 12.0 above retrieval alone. The gain held, again preregistered, for Qwen3-Coder-30B-A3B ($+20.1$ and $+5.4$ points) and Qwen3.8-27B ($+14.1$ and $+13.0$), with all intervals above zero.

\paragraph{Why.}
An agent without evidence never displays 80\% of the package, and finds another 7\% within three commands. The savings track the remaining 12\% that it would otherwise dig for ($\rho=+0.27$). Re-reads of package code are mostly checks of the agent's own edits (40\%) or retrievals after the package was removed (29\%) (\cref{app:value}). Evidence shortens the search for where to edit. It does not supply the judgment of whether an edit is right, the same evidence-to-action gap that trajectory studies report \citep{li2026codetracer}. Every token it adds is paid at every later turn.

%% file: sections/05_removing.tex
\section{Removing Context: The Sign Depends on the Price}
\label{sec:removing}

Observations, the outputs of the agent's own commands, make up most of a coding agent's context. Replacing old observations with a placeholder is a simple and strong baseline for managing it \citep{lindenbauer2025complexity}. The recommended setting keeps the latest ten observations and masks the rest, so from the eleventh command on, every turn changes the prompt at a different point. By \cref{eq:breakeven}, each such change saves the cached price of one old observation on the remaining turns but forces the rest of the prompt to be recomputed. At $c=0.1$ the trade-off should therefore reverse.

\paragraph{Pilot.}
With our agent on 80 Verified tasks, rolling masking cut cost by 41\% at $c=1$ but more than doubled it at $c=0.1$. Masking the oldest ten observations at once whenever twenty had accumulated (\emph{batched}) saved 12\% at $c=0.1$ and 35\% at $c=1$ (\cref{app:robustness}).

\paragraph{Preregistered test.}
We then preregistered four policies in mini-SWE-agent on 201 Verified tasks never used before, with Qwen3.6-27B and Qwen3-Coder-30B-A3B: no masking (\cond{MR}), rolling masking of all but the latest ten (\cond{MM10}), batched masking (\cond{MB}), and break-even masking for $c=0.1$ (\cond{MAR}). \cond{MAR} masks every unmasked observation outside the latest ten whenever \cref{eq:breakeven} holds. It takes $B$ and $S$ from the current prompt and estimates $R=\min(250-t,\ \max(59-t,\ 25))$ from the remaining-step distribution of a pilot. Costs use own-cache hits measured with the tokenizer, and the eight primary tests are Holm-corrected.

\begin{figure}[t]
\centering
\figorbox{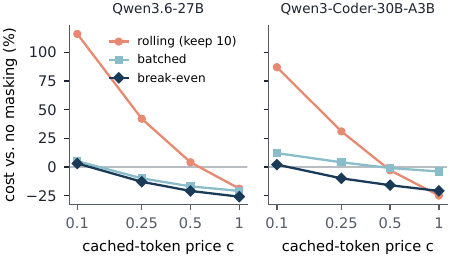}{\linewidth}
\caption{Cost of three masking policies relative to no masking as the price of a cached token varies, on 201 fresh Verified tasks in mini-SWE-agent. Rolling masking doubles cost at $c=0.1$ and saves at $c=1$; break-even masking, set for $c=0.1$, stays near no masking when caching is cheap and saves when it is not.}
\label{fig:price}
\end{figure}

\paragraph{The sign flips.}
Rolling masking raises cost at $c=0.1$ by 116\% for Qwen3.6-27B and by 87\% for Qwen3-Coder. It lowers cost at $c=1$ by 19\% and 25\%. All four tests are supported, and the crossover lies near $c=0.5$ ($+4\%$ and $-3\%$; \cref{fig:price}). Break-even masking costs 52\% and 46\% less than rolling masking at $c=0.1$ (both supported) and stays close to no masking ($+2.8\%$ and $+1.8\%$). Its upper bounds ($+8.3\%$, $+11.4\%$) miss the preregistered non-inferiority margin of 5\%, so we cannot claim it is never worse than not masking. Repriced at $c\ge0.25$, the same runs cost 10--26\% less than no masking. Set for $c=0.025$ (preregistered), it masked 7.4 rather than 36 observations per run and cost 16--19\% less than the rule for $c=0.1$, but only matched no masking ($+0.7\%$, $+0.2\%$). Batched masking behaves similarly for Qwen3.6-27B but saves little for Qwen3-Coder, whose runs are shorter.

\paragraph{The flip holds in a second agent scaffold.}
In SWE-agent, whose history processor masks all but the latest ten observations at every turn (preregistered; Qwen3.6-27B, 115 fresh Django and SymPy tasks), rolling masking raised cost at $c=0.1$ by 101\% and lowered it at $c=1$ by 29\%, both supported, and the same held on 78 tasks from seven other repositories. Break-even masking halved the premium of rolling masking but still cost 9.6\% more than no masking at $c=0.1$: masked runs grew from 92 to 116 steps, a cost that the rule, which counts only the tokens it removes, does not see (\cref{app:removal-more}).

\paragraph{Where the crossover lies.}
A first-order version of \cref{eq:breakeven} for rolling masking predicts crossover prices of 0.19--0.50; the measured ones are 0.38--0.55. It ignores that masked content is re-fetched and runs get longer, and a preregistered test of it on Qwen3.8-27B failed (\cref{app:removal-more}).

\paragraph{The flip holds for other model families.}
With GLM-4.7-Flash in mini-SWE-agent (preregistered; the 115 SWE-agent tasks), rolling masking raised cost at $c=0.1$ by 38\% and lowered it at $c=1$ by 36\%, both supported, with a crossover of 0.21. GLM's runs are long (171 steps on average), so each removal is repaid over many requests, the term $R$ in \cref{eq:breakeven}: break-even masking passed the non-inferiority test at $c=0.1$, costing 12.7\% less than no masking, and it raised local resolution by 7.8 points by avoiding context-limit failures. For Gemma-4-26B-A4B, rolling masking again cost more at $c=0.1$ ($+56\%$, supported), but its saving at $c=1$ was not significant, and break-even masking cost $+14\%$ (\cref{app:removal-more}).

\paragraph{Masking has a second price.}
Masked content is sometimes needed again. For Qwen3.6-27B, rolling masking adds 2.0 re-fetches of masked content per run and break-even masking 1.2. Rolling masking also lengthens its runs by 5 steps. Under the official harness, pooled over both models and 193 tasks, break-even masking is equivalent to no masking in resolution ($+1.3$ points, inside the preregistered $\pm5$ points). Batched masking resolves 3.9 points more, so equivalence is not established, but in the direction of higher, not lower, resolution. The on-target rate drops slightly for Qwen3.6-27B ($-0.04$).

\paragraph{Where the cost of removal comes from.}
To separate the price of a cache break from the agent's reaction, we replayed every unmasked run under each masking rule with its actions held fixed (preregistered; the replay reproduces the logged prompt length and cached prefix for 99.6--100\% of requests, \cref{app:decomp}). The gap between this bill and the unmasked bill is the direct effect of masking; the rest of the gap to the masked runs is the cost of the agent's changed behavior. At $c=0.1$, most of rolling masking's premium is direct in mini-SWE-agent ($+90$ of $+116\%$ for Qwen3.6-27B, $+86$ of $+87\%$ for Coder), and about half of it in SWE-agent ($+52$ of $+101\%$). For Qwen3.6-27B, batched and break-even masking would \emph{save} money with the actions fixed, 3--4\% in mini-SWE-agent and 14--22\% in SWE-agent, all intervals excluding zero (for Coder, 2--3\% more); their premiums fall in the behavioral term of this accounting split, not a causal estimate, which in SWE-agent came with 23--28 extra steps. In mini-SWE-agent it is mostly re-acquisition: for Qwen3.6-27B masking adds 1.5--3.0 reads and 1--2 searches per run, no edits. Little context is safe to drop: only 12\% of observation tokens are provably stale, and masking only those saves 4.6\% at $c=1$ with actions fixed (\cref{app:reaction}). Three further methods (\cref{app:removal-methods}) fit this picture: a controller with reuse probabilities estimated from other runs saved 10\% only at $c=1$; truncating long outputs once, as they enter the context, keeps the cache intact yet cost 4\% more at $c=0.1$ while runs grew from 52 to 60 commands; fine-tuning the agent on its own narrow reads made runs longer and 11\% more expensive \citep[cf.][]{weinberger2026token}. Cost is linear in $c$ and $p$, so a method cheaper at the four corners of the current API range ($c\in\{0.025, 0.1\}$, $p\in\{4,6\}$) is cheaper throughout it; none of the 33 combinations we ran was (\cref{app:decomp,app:pricemap}).

\paragraph{Keeping a digest helps, but does not settle the trade-off.}
Deterministic digests in place of placeholders (\cond{MX}) cost 17.7\% less than batched masking on 78 confirmation tasks but 2.4\% more than no masking, and local non-inferiority failed. They lower the cache price above which masking pays from 0.21 to 0.11, just above the current API range (\cref{app:digest}). For GLM-4.7-Flash, whose runs are long, they cost 13--17\% less than no masking at $c=0.1$ in three preregistered runs, including a confirmation on 78 held-out tasks (bound $-5.0\%$; the first test narrowly failed), breaking even at 0.03; for Qwen3.6-27B in mini-SWE-agent, at 0.13, and for Gemma-4 at 0.09 (no significant saving at $c=0.1$).

%% file: sections/06_reeval.tex
\section{Representative Methods under a Price Vector}
\label{sec:reeval}

Papers on context management compare methods at one implicit price, usually token counts or a single provider's bill. We re-ran six methods in one scaffold, on the same tasks and with the same accounting, and priced every run at several price vectors.

\paragraph{Setup.}
All runs use mini-SWE-agent on the 201 fresh Verified tasks of \cref{sec:removing}, one run per task, method and model, for Qwen3.6-27B, Qwen3-Coder-30B-A3B and Qwen3.8-27B (3{,}618 runs). Besides the four policies above, we ran LLM summarization (\cond{MS}) as in \citet{lindenbauer2025complexity} and AgentDiet-style trajectory reduction (\cond{MD}) \citep{xiao2026reducing}. Both use the agent's own model for their extra calls, which we price like agent requests (\cref{app:reeval}). We preregistered four one-sided tests per model, Holm-corrected: \cond{MS} and \cond{MD} each cost more than \cond{MR} at $c=0.1$ and less at $c=1$.

\begin{table}[t]
\caption{Cost of five context-management methods relative to no management (\cond{MR}), on 201 fresh Verified tasks in mini-SWE-agent, one run per cell, at two prices of a cached token ($p=4$). The last row gives the mean cost of \cond{MR} in thousands of tokens.}
\label{tab:reeval}
\centering
\small
\setlength{\tabcolsep}{3.5pt}
\resizebox{\columnwidth}{!}{%
\begin{tabular}{@{}lrrrrrr@{}}
\toprule
& \multicolumn{2}{c}{Qwen3.6-27B} & \multicolumn{2}{c}{Qwen3-Coder} & \multicolumn{2}{c}{Qwen3.8-27B} \\
\cmidrule(lr){2-3}\cmidrule(lr){4-5}\cmidrule(l){6-7}
Method & $c=0.1$ & $c=1$ & $c=0.1$ & $c=1$ & $c=0.1$ & $c=1$ \\
\midrule
\cond{MM10} rolling masking & $+116\%$ & $-19\%$ & $+87\%$ & $-25\%$ & $+76\%$ & $-24\%$ \\
\cond{MB} batched masking & $+5\%$ & $-22\%$ & $+12\%$ & $-4\%$ & $-14\%$ & $-33\%$ \\
\cond{MAR} break-even & $+3\%$ & $-26\%$ & $+2\%$ & $-21\%$ & $+7\%$ & $-21\%$ \\
\cond{MS} summarization & $+22\%$ & $-14\%$ & $+28\%$ & $+3\%$ & $-19\%$ & $-53\%$ \\
\cond{MD} reflection & $+36\%$ & $+4\%$ & $+31\%$ & $+7\%$ & $+29\%$ & $+2\%$ \\
\midrule
\cond{MR} cost (K) & 149 & 901 & 69 & 347 & 129 & 912 \\
\bottomrule
\end{tabular}}
\end{table}

\paragraph{Summarization can flip too.}
For Qwen3.6-27B, summarization costs 21.6\% more than no management at $c=0.1$ and 14.0\% less at $c=1$. Both tests are supported, so its effect changes sign with the price, as rolling masking does. Each summary rewrites the prompt from the condensed turns on, and the summary calls add 22K uncached tokens per run. For Qwen3-Coder, summarization costs 28.3\% more at $c=0.1$ (supported) and does not save at $c=1$ ($+3.1\%$). For Qwen3.8-27B it is cheaper at both prices ($-19.4\%$, and $-52.6\%$). Whether summarization pays thus depends on the model as well as on the price.

\paragraph{Reflection with the agent's own model does not pay.}
Trajectory reduction costs 36.0\%, 30.7\% and 28.5\% more than no management at $c=0.1$ (supported for the first two models) and saves nothing at $c=1$ ($+3.9\%$, $+6.9\%$, $+2.5\%$). It made 9.3, 3.1 and 5.7 reflection calls per run, and every rewritten output invalidates the cache from that point. Priced at a fifth of the agent's price, the extra calls still leave a premium of 6.5--8.1\% at $c=0.1$. With Qwen3.5-9B actually making the calls (preregistered), the premium was $+2.6\%$, but it rewrote more outputs and lowered the on-target rate (0.587 against 0.622). A reflection module saves only if its calls are much cheaper than the agent's and its rewrites rarely break the cache.

\paragraph{Rankings change with the price.}
The cheapest method at $c=0.1$ is no management for Qwen3.6-27B and Qwen3-Coder, and summarization for Qwen3.8-27B. At $c=1$ it is break-even masking, rolling masking and summarization, respectively. Kendall's $\tau$ between the cost rankings at $c=0.1$ and $c=1$ is $+0.20$, $+0.07$ and $+0.47$. Break-even masking is the only method within 7\% of no management at $c=0.1$ and at least 21\% cheaper at $c=1$ for all three models.

\paragraph{Quality.}
Officially (not preregistered), no method changed resolution detectably for Qwen3.6-27B or Qwen3-Coder, but summarization lowered it by 7.8 points for Qwen3.8-27B, the model on which it was cheapest at both prices. Each cell has one run; prompts, local resolution and caveats are in \cref{app:reeval}. Still, the comparisons that a reader would draw from token counts or from a single price did not hold at another price for two of three models.

%% file: sections/07_related.tex
\section{Related Work}
\label{sec:related}

\paragraph{Coding agents.}
SWE-bench \citep{jimenez2024swe} and its Verified subset \citep{openai2024verified} are the standard tests of issue resolution. Scaffolds range from purpose-built interfaces \citep{yang2024swe} and general platforms \citep{wang2025openhands} to pipelines without an agent loop \citep{xia2024agentless} and bash-only agents \citep{yang2025minisweagent}.

\paragraph{Retrieved context.}
Code graphs, localization agents and repository maps aim to put the right code in front of the model \citep{ouyang2025repograph,chen2025locagent}. Recent work trains exploration sub-agents that return line pointers \citep{zhang2026fastcontext}, builds structural indexes inside the agent \citep{bhola2026code}, or routes through a scout \citep{bhola2026scrouting}. We hold the retriever fixed and measure what its output does to exploration, priced cost and resolution, with length-matched and execution controls.

\paragraph{Context management and caching.}
Observation masking matches LLM summarization for SWE agents \citep{lindenbauer2025complexity}. Other work prunes trajectories with a reflection module \citep{xiao2026reducing}, removes low-value reasoning blocks \citep{wang2026can}, keeps the history units that affect later decisions \citep{dixit2026focus}, compresses memory at every turn with training \citep{hoy2026continuous}, prunes file reads to the lines relevant to a goal \citep{wang2026swe}, pages context on demand \citep{mason2026missing}, and evicts by remaining value in cache-friendly batches \citep{xu2026tokenpilot}. Whether a prefix stays cached also depends on concurrent agents \citep{shao2026efficientagent}, and token reduction need not reduce cost \citep{weinberger2026token}. Studies of agent spending find that input tokens dominate and vary widely between runs \citep{bai2026ai}, and that most input in real use is read from the cache \citep{zhu2026tracelab}. We show in a preregistered test on two models that a recommended masking policy changes sign with the cache price, give a closed-form break-even rule for the trade-off, and identify sibling-condition cache sharing as a bias in cost ablations.

\paragraph{Long and irrelevant context.}
Models use long contexts unevenly \citep{liu2024lost} and can be distracted by irrelevant text \citep{shi2023large}. Our random-code control isolates length from relevance.

%% file: sections/08_discussion.tex
\section{Discussion}
\label{sec:discussion}

\paragraph{Token savings and cost savings should be reported separately.}
Our experiments show that the same context-management method can save money at one cache price and increase cost at another. Reporting total tokens alone hides this difference. Evaluations should report uncached input, cached input and output separately, together with their prices. They should also account for cache sharing between concurrent conditions. Otherwise, one condition may appear cheaper because another condition has already processed its prompt prefix.

\paragraph{Context management changes how agents work.}
Removing observations reduces the amount of information in later prompts, but the agent may compensate by taking more steps. This explains why savings in fixed-action replay do not always carry over to online runs. In SWE-agent, batched and break-even masking would save 14--22\% with actions held fixed. A practical evaluation must also measure what the agent does after information is removed.

\paragraph{The appropriate choice depends on the task and price.}
For the retrieval designs tested here, location pointers are less costly than retaining large evidence packages in the prompt. In localization, appending retrieval results to the agent's answer improves coverage. For removal, none of the fixed configurations we tested was consistently cheaper than retaining the context across the evaluated API price range. In the main Qwen comparisons, batching is cheaper once cached input costs at least a quarter of fresh input, and rolling removal reduces inference cost without a cache discount. The break-even rule is an accounting baseline for this trade-off, not a universally beneficial policy. These choices must also satisfy the required quality constraint. Stopping and context removal affect the same expensive tail of a run, so their interaction is another important question for evaluation.

%% file: sections/09_limitations.tex
\section*{Limitations}
All models are open models served locally, and five of the seven are from the Qwen family. Our settings disable thinking where supported and use bounded output and step budgets; early evidence replications use a 50-step cap that many runs reach. These results do not establish how frontier API agents or longer reasoning budgets would behave. Executable environments cover selected repositories and only tasks whose gold patch passes locally, so they are not representative of every SWE-bench task. The tested evidence package uses one retrieval design and exposure schedule; a missing detectable gain is not proof that retrieval cannot help.

Prices are applied to recorded token streams rather than measured from API invoices. The cache accounting approximates within-run reuse and does not reproduce provider-specific expiration, write charges, breakpoints or load-dependent eviction. We report inference-token cost, not total sandbox compute or wall-clock latency. Break-even masking estimates remaining requests and holds behavior fixed in its calculation; it missed its cost non-inferiority margin for the Qwen models, although it passed for GLM.

Both official and local quality scores use the recorded final working-tree diff of tracked files, not the patch submitted by mini-SWE-agent. An audit of related runs found that staged changes were missed in 1--4\% of runs and that submitted patches differ from working-tree patches in a quarter of runs. We have not recomputed this paper's quality results under a complete base-relative patch definition. Cost accounting is independent of patch scoring, but quality differences of a few points require this qualification. Additional official reports have incomplete coverage and gold self-check failures (\cref{app:officialreturn}); using the official harness on the same patch cannot detect a shared patch-extraction omission.

%% file: sections/10_impact.tex
\section*{Ethical Considerations}
This work studies how adding and removing context change the cost and output quality of LLM coding agents. Its main practical consequence is lower inference cost, which may reduce energy use per task but could also make automated code changes cheaper to produce at scale. All experiments use public benchmarks and open models; no human subjects or private code were involved.

%% file: sections/appendix.tex
\section{Preregistration and Deviations}
\label{app:prereg}
Each experiment was preregistered in a dated research log before it was run: the task set, conditions, number of runs, primary hypotheses, test, multiple-comparison correction and decision rule. Results were written against these rules, including unsupported ones. Failed attempts (for example runs interrupted when a server was restarted) were kept and rerun once; they are excluded from analysis but retained. \Cref{tab:prereg} lists the primary tests reported in this paper.

\begin{table*}[t]
\caption{Preregistered primary tests reported in the paper. One-sided tests use paired bootstrap over tasks; families are Holm-corrected within each experiment.}
\label{tab:prereg}
\centering
\footnotesize
\setlength{\tabcolsep}{4pt}
\resizebox{\textwidth}{!}{%
\begin{tabular}{@{}llll@{}}
\toprule
Experiment & Primary test & Expected & Verdict \\
\midrule
SWE-bench test, Qwen3.6-27B & \cond{E5}$-$\cond{E0} read+search & $<0$ & supported \\
SWE-bench test, Qwen3.6-27B & \cond{EL}$-$\cond{E0} read+search & $<0$ & supported \\
Verified, Qwen3.6-27B & \cond{E5}$-$\cond{E2} prompt tokens & $<0$ & supported \\
Verified, Qwen3.6-27B & \cond{E5}$-$\cond{E0}, \cond{EL}$-$\cond{E0} read+search & $<0$ & supported \\
SWE-bench test, Qwen3-Coder & \cond{E5}$-$\cond{E0} read+search & $<0$ & supported \\
Loc-Bench, Qwen3.6-27B & \cond{E5}$-$\cond{E0} read+search & $<0$ & supported \\
Model ladder (4 models) & slope of on-target gain over size order & $<0$ & not supported \\
Model ladder (4 models) & \cond{E5}$-$\cond{E0} read+search, per model & $<0$ & supported (4/4) \\
Model ladder (4 models) & \cond{E5}$-$\cond{E5r} read+search & $<0$ & supported \\
mini-SWE-agent, Qwen3.6-27B & \cond{E5}$-$\cond{E0} read+search & $<0$ & supported \\
Observation masking & \cond{E0M}$-$\cond{E0} cost at $c=0.1$ & $<0$ & not supported (reversed) \\
Observation masking & \cond{E0B}$-$\cond{E0M} cost at $c=0.1$ & $<0$ & supported \\
Observation masking & interaction with the package, cost & $\neq0$ & not supported \\
Scout and solver & \cond{S}$-$\cond{E0} solver read+search & $<0$ & supported \\
Scout and solver & \cond{S}$-$\cond{E0} total cost (scout $\times9/27$) & $<0$ & not supported \\
Scout and solver & \cond{S}$-$\cond{E0} on target & $\neq0$ & not supported \\
Masking read expansions (Verified + test) & \cond{EDm}$-$\cond{ED} on target, equivalence & within $\pm0.05$ & supported \\
Masking read expansions (Verified) & \cond{EDm}$-$\cond{ED} cost at $c=0.1$ & $<0$ & supported \\
Fine-tuning (Qwen3.5-9B) & three interactions with evidence & see \cref{app:training} & not supported (3/3) \\
Fresh Verified tasks & \cond{E5}$-$\cond{E0} read+search & $<0$ & supported \\
Fresh Verified tasks & \cond{E5}$-$\cond{E0} resolved, equivalence & within $\pm5$\pp & supported \\
Gemma-4-26B-A4B & \cond{E5}$-$\cond{E0} read+search & $<0$ & not supported \\
Removal timing, Gemma-4 & \cond{E5s8}$-$\cond{E0}, \cond{E5src}$-$\cond{E0} read+search & $<0$, $<0$ & not supported (0/2) \\
Pairs of evidence (oracle) & interaction $F{+}G-F-G+\varnothing$, resolved & $\neq0$ & not scored officially \\
Pairs of evidence (oracle) & $F{+}G$ $-$ $F{+}$random, resolved & $>0$ & not scored officially \\
Locations + batched masking & \cond{ELB}$-$\cond{E0} cost, read+search & $<0$, $<0$ & supported (2/2) \\
Locations + batched masking & \cond{ELB}$-$\cond{E0} resolved, equivalence & within $\pm5$\pp & not scored officially \\
\midrule
Break-even masking set for $c=0.025$, 201 tasks, 2 models & vs.\ no masking at $c=0.025$: one-sided upper bound $<+5\%$; vs.\ the rule set for $c=0.1$: $<0$ & all four & 2/4 (vs.\ the $c=0.1$ rule $-18.5\%$, $-16.3\%$; vs.\ no masking $+0.7\%$, $+0.2\%$, bounds $+6.9\%$, $+5.2\%$) \\
Evidence package on Verified, each condition on its own server, 80 tasks & server-reported premium, lower bound $>$ concurrent estimate ($+7.8\%$) & -- & supported ($+28.7\%$) \\
AgentDiet with a 9B reflection model, 201 tasks & cost vs.\ no masking at $c=0.1$ $>0$ & -- & not supported ($+2.6\%$) \\
Masking, 201 fresh tasks, 2 models & \cond{MM10}$-$\cond{MR} cost at $c=0.1$ & $>0$ & supported (2/2) \\
Masking, 201 fresh tasks, 2 models & \cond{MM10}$-$\cond{MR} cost at $c=1$ & $<0$ & supported (2/2) \\
Masking, 201 fresh tasks, 2 models & \cond{MAR}$-$\cond{MM10} cost at $c=0.1$ & $<0$ & supported (2/2) \\
Masking, 201 fresh tasks, 2 models & \cond{MAR}$-$\cond{MR} cost at $c=0.1$, non-inferiority & upper bound $<+5\%$ & not supported (0/2) \\
Masking, 201 fresh tasks, 2 models & \cond{MAR}$-$\cond{MR}, \cond{MB}$-$\cond{MR} resolved, equivalence & within $\pm5$\pp & \cond{MAR} supported ($+1.3$\pp); \cond{MB} not ($+3.9$\pp) \\
Evidence $\times$ execution, 183 tasks & interaction; execution; evidence with execution, resolved & $>0$ & not supported (0/3; $-1.6$, $+3.8$ Holm $p=0.077$, $-0.3$\pp; local: no; yes; no) \\
Localization, 184 Loc-Bench tasks & agent + retrieval $-$ agent; $-$ retrieval, function Acc@10 & $>0$, $>0$ & supported (2/2) \\
Localization, Coder and Qwen3.8-27B, 184 tasks & agent + retrieval $-$ agent; $-$ retrieval, function Acc@10 & $>0$, $>0$ & supported (4/4) \\
\bottomrule
\end{tabular}}
\end{table*}

\paragraph{Deviations.} %
Deviations were recorded in the log when they occurred. Those that bear on results in this paper are:
\begin{itemize}
\item \emph{Candidate rule.} The number of candidate functions ($K=3$) and the rule that skips test files were fixed after seeing offline localization accuracy on the first SWE-bench test batch (top-3 hit rate 65\%, or 52.5\% without skipping test files). This was before any agent run, and the rule was not changed afterwards.
\item \emph{Context overflow.} In the first experiment, three runs failed when the conversation exceeded the 64K context. The agent was changed to end such runs with a separate status, counted as not submitted, and only those runs were repeated with the same seeds.
\item \emph{Truncated expansion files.} The first on-demand batch was stopped after about 40 runs: the 6{,}000-character output limit cut expansion files (about 15{,}000 characters) in the middle, so the agent could not read what \cond{E5} shows in full. Reads from the mounted directory were given a 20{,}000-character limit and the whole batch was rerun; the aborted runs are kept but excluded. No outcome had been examined when this was decided.
\item \emph{Command classification.} Classification first looked at the first word of a command, so a command such as \texttt{cd <path> \&\& grep \dots} counted as other. Tests preregistered after this was noticed strip a leading \texttt{cd} before classifying. For the Qwen models at most 5.4\% of commands start with \texttt{cd}, which changes read and search counts by at most 0.9\%; earlier results are reported as preregistered.
\item \emph{Infrastructure failures.} Runs that failed because a model server on the shared machine was terminated, or because the machine reached its process limit, were set aside (kept, not analyzed) and rerun once with the same task, condition and seed. A run that spans a server restart loses its prefix cache once, which slightly inflates its cost.
\item \emph{Loc-Bench.} The Loc-Bench comparison reuses runs sampled as training data (\cref{app:training}). Its tests were preregistered after the runs existed but before any test was computed. While selecting training data we had seen the number of unsubmitted runs per condition and the mean cost of the selected runs, but not exploration counts or any paired difference, so exploration was chosen as the primary outcome.
\item \emph{Capability order.} The model ladder's preregistered order ranked Qwen3.8-27B strongest; in our harness it was the weakest. We report the preregistered slope test as not supported and treat the order by observed accuracy as exploratory.
\item \emph{Non-Qwen model.} We preregistered GLM-4.7-Flash, but our serving environment could not load it, and we replaced it with Gemma-4-26B-A4B under an amended preregistration written before any Gemma run. Gemma used a newer vLLM version and, by a format rule fixed in advance (more than 10\% of turns without a command in a text-format smoke test), tool calls instead of fenced bash blocks; its cost is therefore compared only within the model.
\item \emph{Cache hits from concurrent conditions.} Preregistered cost tests used server-reported cache hits, which, as we found later, include hits on prefixes computed by a concurrently running sibling condition (\cref{app:cost}). We report costs excluding such hits throughout. Every preregistered cost test keeps its verdict under the corrected measure (for example, \cond{ELB}$-$\cond{E0} $-13\%$ instead of $-18\%$; \cond{EDm}$-$\cond{ED} $-23\%$ instead of $-31\%$), but two descriptive claims changed: the full package's premium on Verified at $c=0.1$ is $+36\%$, not $+8\%$, and batched masking does not bring the full package to cost parity with no evidence.
\item \emph{Executable sandbox.} The SWE-bench specification uses Python 3.6 for Django 3.0--3.2; our installer had no 3.6, so we used 3.8, which these versions support. Dependencies come from each version's environment-setup commit. During environment validation, the shared unprivileged user hit the machine's per-user process limit and 52 masking runs that overlapped this window saw fork failures; they were set aside and rerun with the same seeds, and later sandboxes use a dedicated user.
\item \emph{Cache accounting.} The cache-aware cost analysis is exploratory. Its first version counted the removed tokens themselves as recomputed at the trigger (a median of about 18.6K instead of 4.5K tokens on Verified). We corrected this after reviewing the analysis, and the paper reports the corrected numbers (\cref{app:cost}).
\end{itemize}

\section{Agent Details}
\label{app:agent}
\paragraph{System prompt.}
\begin{quote}\small
You are an expert software engineer fixing a GitHub issue in a Python repository. You interact with the repository only through bash commands. In every reply, think briefly, then give exactly ONE bash command in a single \verb|```bash| code block. The command runs in the repository root; its output is returned to you. Environment limits: no network access; the repository's dependencies are NOT installed, so you cannot run its test suite. Use commands such as grep, find, sed -n, cat, python3 (standard library only) to inspect code, and sed -i, python3 or \verb|cat <<'EOF'| to edit files. Make the minimal correct change. When the fix is complete, reply with the single word SUBMIT and no code block.
\end{quote}

\paragraph{User prompt.} The issue text, followed by the evidence in a \texttt{<context>} block (``The following code from the repository may be relevant. Line numbers are shown on the left.''), followed by ``Fix the issue. Give one bash command per reply; reply SUBMIT when done.'' At the trigger, the evidence block is replaced by a \texttt{<context\_index>} block listing the locations shown, with the instruction to re-read them with \texttt{sed -n} if needed. A reply without a command and without SUBMIT receives a one-line reminder and counts as a turn.

\paragraph{Sandbox.} Each command runs in fresh PID, network, mount, IPC and UTS namespaces as an unprivileged user, with the repository mounted read-write, no network, and limits on CPU time, memory, processes and wall time (90\,s). Output beyond 6{,}000 characters keeps its first and last 3{,}000 characters.

\paragraph{Git in the sandbox.} In our sandboxes the repository belongs to another user, so git refused the agent's own git commands; most runs hit this once or twice (3--5\% of commands). Patches were always taken on the host, and all conditions shared this environment.

\section{Evidence Package Construction}
\label{app:package}
Candidate edit locations come from a dense code-search model applied to the issue text, over functions and methods of the repository at the base commit, skipping test files. The top three candidates are rendered in full with line numbers; the next seven are listed by path, line range and symbol. Related code is selected by a code-graph retriever around the three candidates: definitions they depend on (contracts of called functions and used classes), nearby definitions in the same file, and pointers to further definitions, ranked and truncated so that the whole package fits in 1{,}000 lines. The index that replaces the package lists every location shown. Retrieval never uses the gold patch; gold edit locations are used only to score whether candidates and patches hit the right function. Tasks for which retrieval failed were dropped before running and, in the first Verified set, replaced by a same-repository backup chosen in advance.

\section{Additional Results}
\label{app:tables}
\begin{table*}[t]
\caption{\cond{E5}$-$\cond{E0} on each benchmark and model: paired mean differences. Cost is relative to the \cond{E0} mean. Cost at $c=0.1$ counts only cache hits a run could have produced itself (\cref{app:cost}). SWE-bench test uses the batch of the main end-to-end experiment; \cref{fig:cost} uses a later batch with \cond{EI} and \cond{ED}, where the corresponding costs are $+25\%$ ($c=1$) and $+32\%$ ($c=0.1$).}
\label{tab:app-main}
\centering
\footnotesize
\setlength{\tabcolsep}{4pt}
\resizebox{\textwidth}{!}{%
\begin{tabular}{@{}lrrrrr@{}}
\toprule
Setting & Read+search & Cost, $c=0.1$ & Cost, $c=1$ & On target & Submitted \\
\midrule
SWE-bench test, Qwen3.6-27B & $-5.97$ & $+42\%$ & $+33\%$ & $+0.01$ & $+0.14$ \\
Verified, Qwen3.6-27B & $-6.64$ & $+36\%$ & $+25\%$ & $+0.04$ & $+0.14$ \\
Loc-Bench, Qwen3.6-27B & $-4.52$ & $+36\%$ & $+28\%$ & $+0.02$ & $+0.07$ \\
SWE-bench test, Qwen3-Coder & $-2.97$ & $+95\%$ & $+68\%$ & $+0.03$ & $-0.01$ \\
Verified, Qwen3.5-4B & $-7.54$ & $+25\%$ & -- & $+0.08$ & $+0.19$ \\
Verified, Qwen3.5-9B & $-7.27$ & $+33\%$ & -- & $+0.03$ & $+0.11$ \\
Verified, Qwen3.6-27B (2nd batch) & $-7.01$ & $+33\%$ & -- & $+0.03$ & $+0.14$ \\
Verified, Qwen3.8-27B & $-5.91$ & $+38\%$ & -- & $+0.09$ & $+0.10$ \\
Verified (219 fresh), Qwen3.6-27B & $-4.87$ & $+42\%$ & $+29\%$ & $+0.02$ & $+0.08$ \\
Verified, Qwen3.6-27B, mini-SWE-agent & $-1.29$ & $+27\%$ & -- & $+0.01$ & $+0.06$ \\
\bottomrule
\end{tabular}}
\end{table*}

\paragraph{Locations versus code.} With Qwen3.6-27B, \cond{EL}$-$\cond{E0} read+search is $-3.04$ on SWE-bench test and $-2.02$ on Verified; \cond{E5}$-$\cond{EL} is $-2.94$ and $-4.62$. With Qwen3-Coder, \cond{EL}$-$\cond{E0} is $-2.88$ and \cond{E5}$-$\cond{EL} is $-0.09$.

\paragraph{Cost per resolved task.} %
On SWE-bench test, total cost at $c=0.1$ divided by resolved tasks is $1.59\times$ that of \cond{E0} for \cond{E5}, $2.07\times$ for \cond{E2} and $1.09\times$ for \cond{EL}; on Verified the ratios are $1.26\times$, $1.61\times$ and $0.98\times$.

\section{Evidence Experiments: Figures}
\label{app:evidence-figs}
These figures belong to \cref{sec:adding}; their numbers are summarized there and in \cref{tab:app-main}.

\begin{figure}[h]
\centering
\figorbox{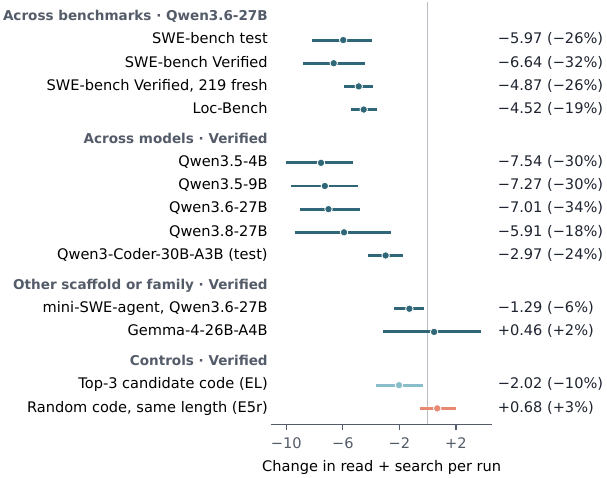}{\linewidth}
\caption{Change in read and search commands per run relative to no evidence (\cond{E0}), with 95\% bootstrap CIs over tasks. The full package (\cond{E5}) reduces exploration on every benchmark and model, though much less or not at all in mini-SWE-agent and with Gemma-4; random code of the same length (\cond{E5r}) does not.}
\label{fig:exploration}
\end{figure}

\begin{figure}[h]
\centering
\figorbox{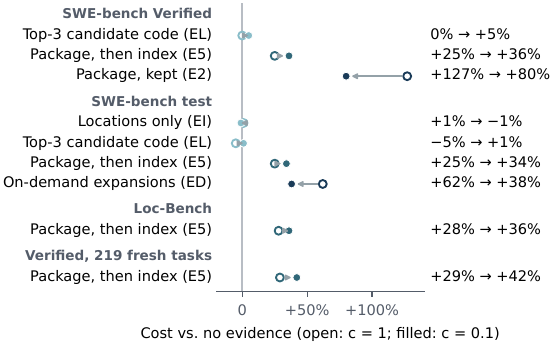}{\linewidth}
\caption{Run cost relative to no evidence for each condition, with cached tokens at full price ($c=1$, open circles) and at 10\% of the price ($c=0.1$, filled), Qwen3.6-27B. Cost follows \cref{eq:cost}, counting only cache hits a run could have produced itself.}
\label{fig:cost}
\end{figure}

\begin{figure}[h]
\centering
\figorbox{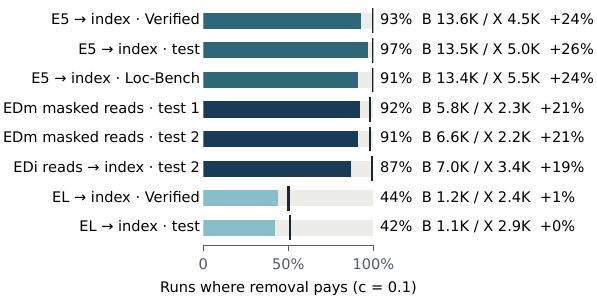}{\linewidth}
\caption{Share of runs in which removing evidence at the trigger pays off under \cref{eq:breakeven} at $c=0.1$, using each run's remaining turns (bars) or the condition's median (ticks). Labels give the median tokens removed ($B$) and recomputed ($X$) and the net saving.}
\label{fig:removal}
\end{figure}

\begin{figure}[h]
\centering
\figorbox{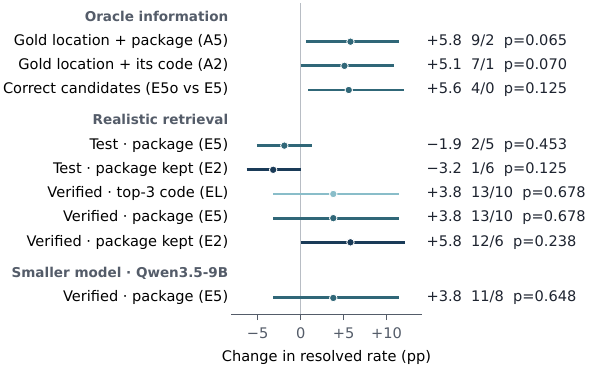}{\linewidth}
\caption{Change in the official resolved rate (percentage points) against each condition's baseline, paired by task, with 95\% bootstrap CIs. Labels give the number of tasks that improved and worsened and an exact two-sided sign test. Tasks whose gold patch fails in our harness are excluded.}
\label{fig:quality}
\end{figure}

\begin{figure}[h]
\centering
\figorbox{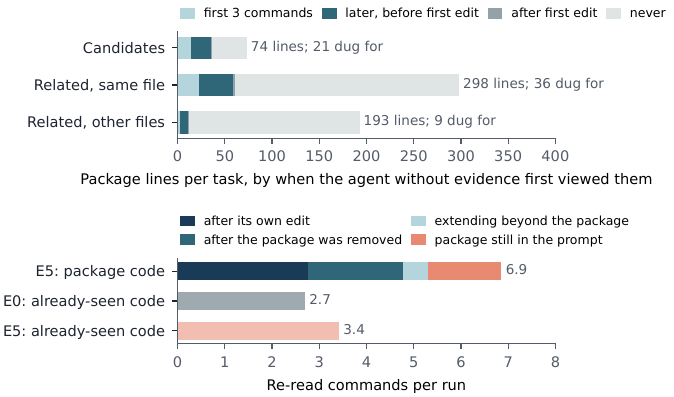}{\linewidth}
\caption{(a) Package lines per task by when the agent \emph{without} evidence first displayed them in the same task. (b) Re-read commands per run: re-reads of package code by the agent \emph{with} evidence, by why the code was not simply in view, and re-reads of code already in context (including the package while it is shown) with and without evidence. Loc-Bench, Qwen3.6-27B.}
\label{fig:value}
\end{figure}

\section{Robustness}
\label{app:robustness}
\paragraph{Fixed-state probe.} %
From 71 states after the first command of a no-evidence run (first command a read or search not naming a gold file), we continued five times with nothing added, with the three candidate locations, with their code, or with the related code (truncated to 6{,}000 characters). The probability that the next command names a gold-edited file was 0.57 with nothing added, 0.74 with locations ($+0.175$), 0.75 with code ($+0.177$) and 0.72 with related code ($+0.149$). This probe was exploratory and its conditions differ in length.

\paragraph{Keeping evidence by code relations.} %
In fixed-state forks at the first edit, we kept the read expansion files whose code relates to the edited function, instead of replacing them with an index or the location list. In a preregistered confirmation on 40 new tasks this did not reduce the remaining exploration (relations versus locations: $+0.45$ commands, relations versus index: $+0.84$), and an exploratory advantage seen earlier did not replicate. We report this negative result because it motivated the incremental-value analysis.

\paragraph{Reranking candidates with the model.} %
Letting the model rerank the candidate list before the run changed neither the on-target rate ($-0.006$) nor exploration ($-0.58$ commands) relative to the original ranking; all three preregistered tests were unsupported.

\paragraph{Other agent framework.} %
We reran \cond{E0} and \cond{E5} with Qwen3.6-27B on the 80 Verified tasks, two runs each, in mini-SWE-agent 2.4.6 \citep{yang2025minisweagent}. We kept its official SWE-bench prompt and loop (one bash command per step in a fenced block, its output format and submission protocol) and changed only the execution environment (our sandbox), the sampling settings (as in our agent) and the step limit (50, as in our agent, instead of the official 250). The package follows the issue text and is replaced by an index with the same lifecycle as \cond{E5}. The preregistered test replicates: the package reduces read and search commands by 1.29 per run ($-6\%$, one-sided $p=0.0095$), raises cost at $c=0.1$ by 27\%, and leaves the on-target rate unchanged ($+0.006$). The reduction is much smaller than in our agent on the same tasks ($-34\%$). Counted over a fixed number of commands, the early saving is similar in both scaffolds: over the first eight commands, $-1.01$ in mini-SWE-agent versus $-1.38$ in our agent. It then stays near one command in mini-SWE-agent ($-0.73$ over the first 24 commands), whereas in our agent it grows ($-3.64$ over the first 24) as runs with the package submit earlier (20.4 versus 28.1 steps). More than half of mini-SWE-agent's runs in each condition reach the 50-step limit (58\% and 52\%); its prompt asks the agent to reproduce the issue, verify the fix and test edge cases, but we do not separate this from task difficulty or the step limit, which is smaller than the official 250. (A first, exploratory split at each condition's own lifecycle trigger overstated the early saving, because runs with the package reach the trigger sooner.)

\paragraph{A model from another family.} %
Gemma-4-26B-A4B ran the same 80 Verified tasks, two runs per condition, served with vLLM 0.26.0 and prompted through tool calls (\cref{app:prereg}). The preregistered reduction in exploration does not replicate ($+0.46$ commands, one-sided $p=0.61$), and cost at $c=0.1$ rises by 27\%. Gemma rarely completes tasks in our harness: 76--79\% of its runs reach the 50-step limit and 62--64\% end with an empty patch. Over the first eight commands the package saves $1.27$ commands (Qwen3.6-27B: $1.38$), but the saving fades later in the run ($-0.82$ over the first 24). Gemma writes a script into the repository within its first two steps, which our lifecycle counts as an edit, so the package is replaced by its index at a median of step 2 (Qwen3.6-27B: step 6). Official resolution was 0.109 without and 0.083 with the package ($-2.6$ points, 78 tasks).

\paragraph{Removal timing.} %
Our lifecycle removes the package at the first edit of any file or at step 8, so the same rule exposes the package for different lengths of time in different models. We preregistered two variants on the same 80 Verified tasks, two runs each: removal at step 8 regardless of edits (\cond{E5s8}), and removal at the first change to a tracked file, at the latest at step 8 (\cond{E5src}; new scratch files do not count). With Gemma-4-26B-A4B the package then stays until a median of step 8 instead of step 2, which extends the early saving (over the first 24 commands, $-2.2$ and $-2.0$ commands, against $-1.0$ under the standard lifecycle), but neither variant reduces exploration significantly over the run ($-1.44$, and $-2.06$, one-sided Holm-adjusted $p=0.28$ for both), and cost at $c=0.1$ remains 29--31\% above no evidence. With Qwen3.6-27B the removal point barely matters: the three lifecycles save 5.9 to 6.9 commands, and neither variant differs from the standard one in exploration or cost ($+0\%$ and $-2\%$). As set out in the preregistration, we did not pursue removal timing further.

\paragraph{Pairs of evidence (oracle).} %
Code from two functions might be useful only together, for example a caller and the callee whose failure behavior it misreads. On 43 Verified tasks where the gold-edited function $F$ relies on another function $G$ named in the issue and called in the fix, we showed nothing, $F$ alone, $G$ alone, both, or $F$ with a random function of $G$'s length, four runs each, rendered from the base commit (an oracle manipulation like that of \cref{sec:adding}; the patch is never shown). We found no combination benefit: the interaction on the on-target rate is $+0.023$, and given $F$, adding $G$ is no better than adding random code ($-0.012$). In exploration the two halves substitute for each other: $F$ alone saves 4.8 commands, $G$ alone 2.8, and both together 4.6 (interaction $+3.0$). Without evidence the agent reads $G$ itself in 43\% of runs. Official resolution was not scored for this variant.

\paragraph{Masking read expansions.} %
In the on-demand condition, the agent reads expansion files from a mounted directory. Replacing the expansion files it has already read with a one-line note at the trigger (\cond{EDm}) is equivalent to keeping them in edit accuracy: pooled over 80 Verified and 80 SWE-bench test tasks, the on-target difference is $+0.005$ within the preregistered $\pm0.05$ margin. On Verified it cuts cost at $c=0.1$ by 23\% ($-17.6$K tokens per run; 31\% under the preregistered measure with server-reported hits, \cref{app:cost}), and the agent does not re-read the removed files more often afterwards (0.09 versus 0.11 re-reads per run). Official resolution was not scored for this variant.

\paragraph{Observation masking.} %
Following \citet{lindenbauer2025complexity}, we replaced old command outputs with a one-line placeholder, either keeping only the latest ten (rolling, \cond{E0M}) or, once twenty had accumulated, masking the oldest ten at once (batched, \cond{E0B}); \cond{E5M} applies rolling masking with the package. Without a cache discount, rolling masking cuts cost by 41\% ($-112$K tokens per run). But it rewrites the prompt prefix on every turn: the cache hit rate falls from 82\% to 38\%, and at $c=0.1$ cost more than doubles ($+65.6$K, $+111\%$), so the preregistered saving is not supported. Batched masking breaks the cache only once every ten turns, keeps the hit rate at 82\%, and costs less than rolling masking ($-74.4$K, supported) and less than no masking at either price ($-12\%$ at $c=0.1$, $-35\%$ at $c=1$), while the on-target rate rises slightly ($+0.05$). Masking and the package do not interact in cost (difference in differences $-1.8$K); with the package, rolling masking lowered the on-target rate ($-0.05$). Official resolution was not scored for this variant.

\paragraph{Localization only.} %
Following LocAgent \citep{chen2025locagent}, we asked the agent (mini-SWE-agent template with the task and submission changed, read-only, 50 steps) to return up to ten ranked locations instead of a patch, scored by file Acc@1/3/5 and function Acc@5/10. On 40 Loc-Bench tasks, one run per cell, file Acc@1 was 0.750 without evidence, 0.675 with the location index, 0.625 with the three candidate functions and 0.600 with the full package. The differences came mostly from runs that did not submit within 50 steps (2, 4, 6 and 7 runs); on the 30 tasks where all four submitted, the conditions were level (function Acc@10 0.50--0.57). The retrieval ranking alone reached file Acc@5 0.80 and function Acc@10 0.55. Because the agent submits only one or two locations, appending the retrieval ranking to its answer raised function Acc@10 from 0.50 to 0.675 at no extra cost. This pilot is descriptive. On the 184 held-out tasks (\cref{sec:adding}), appending retrieval helped all three agents. Part of this gain only fills the slots the agent leaves empty: appending instead the functions the agent itself saw in its run, in order of first appearance, raised function Acc@10 by 0.09--0.14 (descriptive). Retrieval added 0.065 and 0.060 beyond that for Qwen3.6-27B and Qwen3-Coder, and 0.016 for Qwen3.8-27B. In an exploratory comparison, merging the answers of two or three agents before the retrieval ranking added only 1--3 points (function Acc@10 0.549--0.571) for two to four times the cost. Qwen3.8-27B with retrieval reached 0.549 at 19.9K tokens per task, against 0.538 at 45.3K for Qwen3.6-27B with retrieval.

\paragraph{A unified context controller.} %
The controller scored each block (observation, evidence snippet, or the payload of an old edit command) by its size and a reuse probability: a prior by command type (read 0.5, search 0.3, script 0.2, edit 0.1, other 0.05), set to zero when the block's source file had changed since it was read (write-invalidate, detected from content hashes) and to one when the agent re-requested masked content (pinning). It removed, at the cut point with the largest net value under \cref{eq:breakeven}, all blocks after it whose expected saving exceeded their expected re-fetch cost. Three versions (the last with priors calibrated to the observed extra re-fetches, about 0.1 per block) ran on 24 Django and SymPy tasks, one run per cell. None beat break-even masking: the third version cost 6\% more at $c=0.1$ and 25\% more at $c=1$. Asking the model to keep its reasoning to two sentences saved 13--23\% in two rounds without lowering local resolution, mostly through shorter runs; a preregistered test of this prompt belongs to companion work on stopping.

\paragraph{Scout and solver.} %
Instead of the code-graph package, a Qwen3.5-9B scout explored read-only for up to 15 commands and handed over one to five locations with a short note; we verified each location against the repository, rendered its code like the package, and gave it to the Qwen3.6-27B solver with the same lifecycle (\cond{S}). Every scout produced a handoff, with 1.1 valid locations on average, containing a gold-edited function in 57\% of runs. The handoff reduced the solver's exploration about as much as the package ($-5.98$ commands relative to \cond{E0}, supported) and cut the solver's own cost by 20\%. Counting the scout at 9/27 of its cost, the total saving ($-8\%$) was not significant; counted in full, the total rose by 15\%. Edit accuracy did not change significantly relative to no evidence ($-0.025$) but was lower than with the full package ($-0.10$); on the 25 tasks where the handoff missed the gold function it fell by 0.12 (exploratory). Official resolution was not scored for this variant.

\paragraph{Execution and localization.}
In the crossed experiment of \cref{sec:adding}, execution alone raised official resolution by 3.8 points (one-sided $p=0.026$), which misses significance after Holm correction ($p=0.077$), and the package cut exploration by 2.2 commands. On localization alone, scored by Acc@$k$ as in LocAgent \citep{chen2025locagent}, the package did not improve the agent's ranked locations in a 40-task pilot.

\section{Three Further Removal Methods}
\label{app:removal-methods}
We preregistered three methods that avoid rewriting old context or learn what to keep, each run twice on 178 Django and SymPy tasks with executable environments and paired with two unmodified runs of Qwen3.6-27B under the same seeds; their parameters and training data come only from other tasks. \emph{Truncation} cuts every command output longer than 2{,}000 characters to its first 1{,}400 and last 400 characters with a note, once, when it enters the context. \emph{Narrow reads} fine-tunes the agent with LoRA (one epoch) on 1{,}118 of its own steps that read with \texttt{sed -n}, \texttt{grep -n}, \texttt{head} or \texttt{tail} and returned at most 1{,}000 characters, plus 460 ordinary steps. \emph{Estimated reuse} is the controller of \cref{sec:removing} with the probability that a block is used again estimated per command type from other runs (reads 0.56, searches 0.25). A method had to save at both prices (Holm over six tests) and keep resolution within 5 points.

\begin{table*}[t]
\caption{Further removal methods on 178 tasks, two runs each. Cost relative to no removal; local resolution.}
\label{tab:removal-methods}
\centering
\footnotesize
\setlength{\tabcolsep}{3pt}
\begin{tabular}{@{}lrrrr@{}}
\toprule
Method & Commands & Cost, $c=0.1$ & Cost, $c=1$ & Resolved \\
\midrule
None & 51.9 & -- & -- & 0.632 \\
Truncation & 60.1 & $+3.6\%$ & $+5.6\%$ & 0.612 \\
Narrow reads & 57.9 & $+11.2\%$ & $+16.3\%$ & 0.649 \\
Estimated reuse & 53.4 & $-0.7\%$ & $-10.2\%$ & 0.610 \\
\bottomrule
\end{tabular}
\end{table*}

Only the saving of estimated reuse at $c=1$ was supported, and only narrow reads kept resolution within the margin ($+1.7$ points). Truncation fired 9.5 times per run, and runs grew by about eight commands.

\section{Deterministic Digests with Retrieval of Original Outputs}
\label{app:digest}
\cond{MX} uses the same keep-ten, polling-ten schedule as batched masking, but replaces each omitted observation with a deterministic digest. It retains selected definition lines, search hits, paths, test outcomes and edit confirmations, leaves outputs of at most 300 characters intact, and advertises a \texttt{recall k} command that returns the original output of step $k$ without executing that command again. Digests are fixed after creation. No model generates the digest and no hidden-test information enters it.

After a development pilot on 115 tasks, the extraction rules were left unchanged for the preregistered confirmation on 78 tasks from seven other repositories. These tasks were held out from this method's pilot, not unused throughout the project. \cond{MX} and no masking each have two runs per task; batched masking has one, so comparisons against it use only repeat zero. Bootstrap resampling uses tasks after averaging repeats within task (2,000 draws). The two primary cost tests use Holm correction; the quality test requires a 95\% lower bound above $-5$ percentage points. All 156 \cond{MX} runs completed and local scoring coverage was checked before interpretation.

At $(c,p)=(0.1,4)$, \cond{MX} costs 17.7\% less than batched masking, supporting that primary test, but 2.4\% more than no masking, failing the other. Local resolution is 0.673 against 0.692 without masking, a difference of $-1.9$ points; non-inferiority is not established. These estimates use local executable-sandbox scores. Returned official comparisons on their matched subsets are reported separately in \cref{app:officialreturn} and do not replace these local denominators. No run used \texttt{recall}; this study therefore does not establish the benefit of retrieval itself. Relative to no masking, the cost difference is $+32.9\%$ at $c=0.025$ and $-23.1\%$ at $c=1$, with $p=4$.

Fixed-action replay of all 156 unmasked trajectories reproduces all 13,330 prompt lengths and own-cache prefixes, with no excluded runs. At $(c,p)=(0.1,4)$, the direct accounting component is $-17.4\%$ and the residual is $+19.8\%$, summing to the observed $+2.4\%$. For a like-for-like repeat-zero comparison, the residual falls from $+43.4\%$ with batched placeholders to $+17.4\%$ with digests. This is an accounting decomposition, not an identified causal mediation effect. At $c=0.025$, the direct component is already positive ($+11.5\%$): reducing re-reading alone need not make this schedule economical.

\paragraph{Break-even cache prices.} %
Cost differences are linear in $c$, so the price at which a method breaks even with no masking follows from its differences at $c=0.1$ and $c=1$. On the 115 development tasks (one run per task), it is 0.224 for batched placeholders, 0.121 for digests, 0.125 for price-aware digests, and does not exist for exact repeated-block references, which cost more at every price ($+4.0\%$ at $c=0.1$, $+4.4\%$ at $c=1$): they keep the cache intact but lengthen runs (92.1 to 98.1 commands). On the 78 confirmation tasks the break-even prices are 0.208 for batched placeholders and 0.113 for digests. These are descriptive; the confirmation's preregistered tests concern $c=0.1$ only.

\paragraph{Digests in mini-SWE-agent.} %
We applied the same digests, without \texttt{recall}, to mini-SWE-agent through a separate wrapper that keeps the batched schedule and changes only the placeholder text (preregistered). For GLM-4.7-Flash on the 115 tasks of \cref{sec:removing}, paired with existing unmasked and batched runs under the same seeds, digests cost 12.7\% less than no masking at $c=0.1$ (Holm $p=0.14$, not supported) and 3.2\% less than batched placeholders (not supported). A second run per task, specified before the first was analyzed, gave $-16.7\%$ (one-sided bound $-2.0\%$, $p=0.03$, supported); pooled, $-14.8\%$ at $c=0.1$, $+4.2\%$ at $c=0.025$ and $-26.5\%$ at $c=1$, a break-even price of 0.033. Context-limit failures fell from 40 to 18 of 230 runs and local resolution changed by $+2.6$ points. A preregistered confirmation on the 78 new-repository tasks, which GLM had not run before (two runs per task, unmasked and digest conditions run at the same time on separate servers), gave $-16.6\%$ at $c=0.1$ (one-sided bound $-5.0\%$, $p=0.008$, supported), $+7.2\%$ at $c=0.025$ and $-30.5\%$ at $c=1$, a break-even price of 0.036. Local resolution changed by $-3.2$ points, so non-inferiority within 5 points is not established there. For Qwen3.6-27B on 178 further tasks with test execution (two runs each), digests cost 4.3\% more than no masking at $c=0.1$ (not supported) and 18.1\% less at $c=1$, a break-even price of 0.134, with local resolution non-inferior ($+0.6$ points). For Gemma-4-26B-A4B on the 115 tasks (one run per task, preregistered together with batched placeholders, which Gemma had not run), digests cost 1.4\% less than no masking at $c=0.1$ (not supported) and 7.9\% less than batched placeholders (not supported), $+12.3\%$ at $c=0.025$ and $-12.9\%$ at $c=1$, a break-even price of 0.086. Batched placeholders cost 26\% more than no masking at $c=0.025$. Context-limit failures fell from 22 to 10 runs; local resolution was 0.342 in both conditions, with an interval too wide ($\pm9$ points) to establish non-inferiority.

\paragraph{Price-aware digest pilot.}
A subsequent preregistered development pilot combined deterministic digests with a price-aware break-even trigger on the same 115 development tasks (one run per task). It retained the problem statement and ten recent observations, and replaced eligible older outputs only when estimated future cached-token savings exceeded the immediate suffix cache invalidation cost. At $(c,p)=(0.1,4)$, cost increased by $4.4\%$ relative to unmasked context (descriptive pilot estimate). Local sandbox resolution was $94/115$ versus $92/115$, a difference of $1.7$ percentage points. All 12,618 prompt-token counts matched the tokenizer audit. The prespecified advancement rule required a negative cost point estimate as well as a quality floor; the cost condition failed, so no confirmation runs were added. These development results are not independent confirmation or official benchmark scores.
Relative to the original digest policy on these same 115 tasks, the price-aware variant changed cost at $(c,p)=(0.1,4)$ by $+0.94\%$ (descriptive pilot estimate). Both groups had complete local scoring; they resolved 94 and 93 tasks, respectively. The new variant produced 5,802 summaries and no recall calls.

\paragraph{Exact repeated-block references.}
A further preregistered development pilot replaced a new observation's exactly repeated blocks (at least six consecutive lines and 400 characters) with references to earlier, fully retained observations. Nonmatching text and the original source remained available; compressed observations could not become reference sources, and existing history was not rewritten. Fixed-action development replay suggested a 4.47\% cost reduction at $(c,p)=(0.1,4)$; one trajectory had a one-token rendering discrepancy on 23 requests, and a sensitivity calculation retaining all tasks and assigning that task zero savings still gave 4.46\%. This replay is not an estimate of realized agent benefit.
In live runs on the same 115 development tasks, cost instead increased by 4.0\% relative to unmasked context (descriptive pilot estimate). Mean commands increased from 92.1 to 98.1. Local resolution was $93/115$ versus $92/115$, a difference of one resolved task. All 115 runs and local scores were present; the audit covered 11,299 requests and 2,060 references. The prespecified negative-cost advancement condition failed, so no confirmation was added. These results do not establish a causal explanation for the additional actions and are not official benchmark scores.

\section{Details of the Removal Experiments}
\label{app:removal-more}

\paragraph{SWE-agent.}
We repeated the comparison in SWE-agent, whose built-in history processor masks all but the latest ten observations at every turn (preregistered; Qwen3.6-27B, 115 other fresh Django and SymPy tasks in the executable sandbox, one run per task, paired with unmasked runs under the same seeds). Rolling masking raised cost at $c=0.1$ by 101\% and lowered it at $c=1$ by 29\%, both supported; the crossover lies near $c=0.40$. On 78 tasks from seven other repositories the same comparison gave $+111\%$ and $-29\%$. As in mini-SWE-agent, masked runs were longer (116 against 92 steps): they re-ran 5.7 more commands whose output had been masked, against 2.0 in mini-SWE-agent. Local resolution did not change detectably ($-0.9$ points). Break-even masking, with $R$ estimated the same way from separate SWE-agent runs (preregistered), cost 46\% less than rolling masking at $c=0.1$ and 25\% less than no masking at $c=1$. At $c=0.1$, however, it cost 9.6\% more than no masking, against $+2.8\%$ in mini-SWE-agent; on the 78 other tasks the three numbers were $-46\%$, $-24\%$ and $+14\%$. It masked 19 times per run, and masked runs grew from 92 to 116 steps, a cost that the rule, which counts only the tokens it removes, does not see. A variant that also charges each masked observation its expected share of the extra steps (estimated on these 115 tasks, preregistered, tested on the other 78) masked a third as often and cut the premium to $+4.3\%$, but its interval still exceeds the 5\% margin, and in mini-SWE-agent it did not help ($+7.0\%$). Batched-control results across prices are in \cref{app:decomp}.

\paragraph{Other model families.}
All results above use Qwen models. We repeated rolling and break-even masking with GLM-4.7-Flash in mini-SWE-agent with tool calls (preregistered; the 115 tasks of the SWE-agent comparison, one run per task, paired with unmasked runs under the same seeds, token counts taken from the server). GLM's runs are long: 171 steps on average, and 21 of the 115 unmasked runs end at the 131K-token context limit. Rolling masking raised cost at $c=0.1$ by 38\% and lowered it at $c=1$ by 36\%, both supported. The crossover, 0.21, is lower than for any Qwen model, as \cref{eq:crossover} predicts for longer runs, although the estimate (0.15) is again too low. With $R$ estimated from separate GLM runs ($L=167$), break-even masking passed the non-inferiority test at $c=0.1$, which it had missed for the Qwen models: it cost 12.7\% less than no masking, 37\% less than rolling masking, and 38\% less than no masking at $c=1$. When many requests remain, each removal is repaid many times over, which is the term $R$ in \cref{eq:breakeven}. Masking also cut context-limit failures from 21 to 4--6 runs and raised local resolution, by 3.5 points for rolling and 7.8 for break-even masking. That is an effect on the agent's behavior, not on pricing. With batched masking added (descriptive; $-10\%$ at $c=0.1$, $-26\%$ at $c=1$), break-even masking was the cheapest of the four policies at every price for GLM, whereas for Qwen3.6-27B in SWE-agent the cheapest policy changed with the price. Which policy is best depends on the price and on how long the agent's runs are, and the break-even rule takes both into account. For a third family, Gemma-4-26B-A4B (preregistered, same design), rolling masking again cost more at $c=0.1$ ($+56\%$, supported) and less at $c=1$ ($-19\%$), but the saving at $c=1$ was not significant: Gemma's runs vary widely in cost, many ending early on format errors and others at the context limit. Its crossover was 0.36 (estimate 0.19). Break-even masking cost 27\% less than rolling masking at $c=0.1$ but $+14\%$ relative to no masking, with too wide an interval to show non-inferiority; Gemma's runs are shorter than GLM's (103 against 171 steps).

\paragraph{Re-evaluation setup.}\label{app:reeval}
All runs use mini-SWE-agent on the 201 fresh Verified tasks of \cref{sec:removing}, one run per task, method and model, for Qwen3.6-27B, Qwen3-Coder-30B-A3B and Qwen3.8-27B (3{,}618 runs without infrastructure errors). The methods are no management (\cond{MR}), rolling masking of all but the latest ten observations \citep{lindenbauer2025complexity} (\cond{MM10}), batched masking (\cond{MB}, close in spirit to batched eviction \citep{xu2026tokenpilot}), break-even masking (\cond{MAR}), LLM summarization (\cond{MS}) and trajectory reduction by a reflection module (\cond{MD}). \cond{MS} follows the LLM-summary baseline of \citet{lindenbauer2025complexity}: once 31 turns have accumulated, it condenses the previous summary and the older turns with the OpenHands summarization prompt, keeping the latest ten. \cond{MD} follows AgentDiet \citep{xiao2026reducing} with $a=2$, $b=1$ and $\theta=500$, rewriting command outputs only. Both use the agent's own model at temperature 0 for their extra calls, which we log and price like agent requests. Because AgentDiet uses a cheaper model for reflection, we also report the extra calls at a fifth of the price. We preregistered four one-sided tests per model, Holm-corrected: \cond{MS} and \cond{MD} each cost more than \cond{MR} at $c=0.1$ and less at $c=1$.

\paragraph{Reversing pairs.}
Three, four and zero pairs of methods (Qwen3.6-27B, Qwen3-Coder, Qwen3.8-27B) reverse their cost order between $c=0.1$ and $c=1$ with both differences outside zero, and every reversing pair involves rolling masking or no management.

\paragraph{Re-evaluation quality and caveats.}
Local resolution on the 115 tasks with an executable environment varies by method between 0.61 and 0.71 for Qwen3.6-27B, 0.26 and 0.32 for Qwen3-Coder, and 0.66 and 0.75 for Qwen3.8-27B. Officially (not preregistered), no method changed resolution detectably for Qwen3.6-27B or Qwen3-Coder, but summarization lowered it by 7.8 points for Qwen3.8-27B, the model on which it was cheapest at both prices. Our summarization prompt is the published OpenHands prompt rather than the modified one of \citet{lindenbauer2025complexity}, which is not public. AgentDiet's reflection prompt is not public either, so we wrote our own from its description. Each cell has one run, and all runs use one scaffold. Still, the comparisons that a reader would draw from token counts or from a single price did not hold at another price for two of three models.

\paragraph{A first-order crossover.}
Rolling masking removes one observation of about $o$ tokens per turn and keeps the latest $K$ turns. Each removal therefore forces about $S\approx K(o+a)$ tokens to be recomputed, where $a$ is the length of a reply. By \cref{eq:breakeven} a removal pays when $c\,o\,R > (1-c)\,K(o+a)$. Averaged over removals, a first-order estimate of the crossover price is
\begin{equation}
c^\ast \approx \Bigl(1 + \frac{\bar R}{K\,(1 + a/o)}\Bigr)^{-1},
\label{eq:crossover}
\end{equation}
where $\bar R$ is the mean number of requests that remain when an observation is removed. The unmasked runs give $c^\ast\approx0.36$ for Qwen3.6-27B, $0.50$ for Qwen3-Coder and $0.19$ for Qwen3.8-27B. Because cost is linear in $c$, the measured crossovers follow exactly from two prices: 0.55, 0.47 and 0.38. The estimate is right for one model and too low for the other two, and a preregistered test of it on Qwen3.8-27B failed: it ignores that masked content is re-fetched and runs get longer, which raises the crossover. The rule explains why rolling masking fails at cheap caches, but not, by itself, where it starts to pay.

\section{Decomposing the Cost of Masking}
\label{app:decomp}

\paragraph{Batched control across prices.}
Batched masking, SWE-agent's polling option that masks ten observations at a time (descriptive), cost 24\% more than no masking at $c=0.1$ and 21\% less at $c=1$, and break-even masking was cheaper than batched masking at all four prices we evaluated. No policy was cheapest at every price: no masking won at $c=0.1$, break-even masking at $c=0.25$ and $0.5$, and rolling masking at $c=1$, on both task sets.

\begin{figure}[h]
\centering
\figorbox{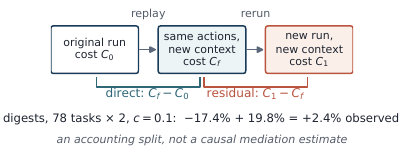}{\linewidth}
\caption{Fixed-action replay. Replaying the unmasked run's actions under the masking rule gives the direct component; the rest of the observed difference is the residual. The example is the digest confirmation of \cref{app:digest}.}
\label{fig:decomp}
\end{figure}

For each unmasked run we rebuilt the prompt of every request as the masking rule would have sent it, with the agent's messages and outputs unchanged, and tokenized it with the serving model's tokenizer and chat template (with the tool definitions in SWE-agent). The rules are those of the runs: the same keep-ten window, batch size, placeholder text and, for break-even masking, the same parameters and token counts. Replaying the unmasked rule itself reproduced the logged prompt length and own-cache prefix for 10{,}538 of 10{,}577 requests (Qwen3.6-27B), 6{,}705 of 6{,}705 (Coder), 10{,}587 of 10{,}610 and 6{,}645 of 6{,}645 (SWE-agent, two task sets). Output tokens are held at their logged values. $K(g,\tau)$ denotes the cost of trajectory $\tau$ rendered under rule $g$; with $\tau_0$ the unmasked and $\tau_1$ the masked run of a task, the direct effect is $K(1,\tau_0)-K(0,\tau_0)$ and the behavioral effect $K(1,\tau_1)-K(1,\tau_0)$. This is an accounting identity, not a causal mediation analysis: a masked agent would not have taken the same actions.

\begin{table*}[t]
\caption{Masking cost relative to no masking at $c=0.1$, split into the direct effect (actions fixed) and the agent's reaction. mini: 201 tasks; SWE-agent: 115 and 78 tasks.}
\label{tab:decomp}
\centering
\footnotesize
\setlength{\tabcolsep}{3pt}
\begin{tabular}{@{}llrrrr@{}}
\toprule
Setting & Rule & Total & Direct & Behavior & Steps \\
\midrule
mini, Qwen3.6-27B & Rolling & $+116\%$ & $+90\%$ & $+26\%$ & $+5.3$ \\
 & Batched & $+4.9\%$ & $-3.4\%$ & $+8.2\%$ & $+2.9$ \\
 & Break-even & $+2.8\%$ & $-3.7\%$ & $+6.5\%$ & $+2.7$ \\
mini, Coder & Rolling & $+87\%$ & $+86\%$ & $+1.2\%$ & $-0.1$ \\
 & Batched & $+12\%$ & $+2.2\%$ & $+9.8\%$ & $+1.8$ \\
 & Break-even & $+1.8\%$ & $+3.4\%$ & $-1.6\%$ & $-0.4$ \\
SWE-agent, 115 & Rolling & $+101\%$ & $+52\%$ & $+49\%$ & $+23.7$ \\
 & Batched & $+24\%$ & $-21\%$ & $+45\%$ & $+23.2$ \\
 & Break-even & $+9.6\%$ & $-15\%$ & $+25\%$ & $+23.4$ \\
SWE-agent, 78 & Rolling & $+111\%$ & $+59\%$ & $+52\%$ & $+28.2$ \\
 & Batched & $+22\%$ & $-22\%$ & $+43\%$ & $+26.4$ \\
 & Break-even & $+14\%$ & $-14\%$ & $+28\%$ & $+27.8$ \\
 & + step charge & $+4.3\%$ & $-18\%$ & $+22\%$ & $+15.9$ \\
\bottomrule
\end{tabular}
\end{table*}

\paragraph{The current price range.} Cost differences are linear in $c$ and $p$, so a method is cheaper than no management throughout $0.025\le c\le0.1$, $4\le p\le6$ if and only if it is cheaper at the four corners. We tested this (preregistered; one-sided 95\% bounds over tasks) for every method and setting we ran: rolling, batched and break-even masking, LLM summarization and AgentDiet-style reduction for three Qwen models in mini-SWE-agent; break-even masking with and without the step charge, truncation, estimated reuse and narrow reads with test execution; rolling, batched and break-even masking in SWE-agent on both task sets; and the same masking rules for GLM-4.7-Flash and Gemma-4. None of the 33 combinations, including batched masking for Gemma-4, was cheaper at all four corners. At $c=0.1$ a few were cheaper in point estimate but not significantly: break-even and batched masking for GLM ($-12.7\%$ and $-9.8\%$), summarization and batched masking for Qwen3.8-27B ($-19\%$ and $-14\%$), and estimated reuse ($-1.7\%$). All of these were more expensive at $c=0.025$. Break-even masking was set for $c=0.1$; at $c=0.025$ its own rule would mask less often, which we did not run.

\section{Cost Accounting}
\label{app:cost}
Cost follows \cref{eq:cost} with the cached-token counts reported by the server for every request, under eight concurrent runs per server, with one correction. Conditions of the same task and repeat share a seed and are submitted together, so a condition whose prompt is identical to a sibling's up to some point (\cond{E5} and \cond{E2} before removal, \cond{EL} and \cond{ELB} before masking starts, \cond{ED} and \cond{EDm} before the trigger) can hit the sibling's prefix cache. We count as cached only the tokens that the run's own previous request could have supplied: hits on the first request, and hits beyond the length of the previous request's prompt, are priced as uncached. This affects 3.7--6.7\% of prompt tokens in conditions that trail an identical sibling (\cond{E5} behind \cond{E2}, \cond{E5B}, \cond{E5M}), about 1--3\% in \cond{EDm}, \cond{EDi}, \cond{ELB} and \cond{E0B}, and none in \cond{E0}. It leaves cost at $c=1$ unchanged. Preregistered cost tests were specified on server-reported hits; we report their verdicts under both measures (\cref{app:prereg}). A cache break is a request whose cached tokens fall below half of the previous request's prompt. For the break-even rule we take, at the request after removal, $X$ as its uncached tokens minus the median uncached tokens of the run's ordinary requests (the earlier estimate, previous prompt minus current cache hits, included the removed tokens themselves and overstated $X$ about fourfold); $B$ as the removed tokens minus the tokens of the index that replaces them; and $R$ as the remaining requests, either actual or, for the ex-ante rule, the condition's median. We also report costs with a 25\% surcharge on cache writes; conclusions do not change.

\paragraph{Audit of the two measures.} %
On the 201-task masking runs of \cref{sec:removing}, where we recorded both, we compared the token-level count with the server's. For Qwen3.6-27B, server hits never exceeded the previous prompt in runs without concurrent siblings (0 of 27{,}389 requests in two later run sets) and did so in 3.6\% of requests with three siblings, all attributable to them; the previous prompt was never an exact prefix of the next, because the chat template re-renders the history, so the previous reply cannot be reused. For Qwen3-Coder, which keeps the history verbatim, server hits exceeded the previous prompt in 83\% of requests: its server also reused the previous reply. Counting that reply as cached when the previous prompt is a prefix moves rolling masking at $c=0.1$ from $+86.7\%$ to $+98.3\%$ over no masking, batched masking from $+12.0\%$ to $+12.9\%$ and break-even masking from $+1.8\%$ to $+2.9\%$, and leaves $c=1$ unchanged. The earlier measure $\min(\hat H_t, P_{t-1})$ counts 9.8\% fewer hits than the token count for Qwen3.6-27B, mostly because the server reports hits in 784-token blocks; under it, rolling masking costs $+173\%$ instead of $+116\%$ at $c=0.1$. It is used only where token ids were not recorded, all in runs that never mask the history. There the true own-cache hit lies between this measure and the whole previous prompt (except at the request that replaces the package). Under that upper bound the package's premium at $c=0.1$ in \cref{tab:app-main} grows from $+25$--$42\%$ to $+31$--$56\%$ for the Qwen3.5 to Qwen3.8 models and stays at $+95\%$ for Qwen3-Coder, so the reported premiums are conservative.

\section{Incremental-Value Analysis}
\label{app:value}
We use the 1{,}792 Loc-Bench runs whose full transcripts were saved. A package line counts if, after stripping whitespace, it has at least six characters and occurs once in its file at the base commit; 59\% of package lines qualify. For each command we match its actual output, after removing line-number prefixes, against the countable lines of the files named in the command and of the paths prefixed in \texttt{grep} output. An item of the package is \emph{viewed} once at least half of its countable lines have been displayed; results with ``at least one line'' as the criterion, and with one or five commands as the boundary for ``quickly'', change the never-or-quickly share between 81\% and 92\%. The correlation between dug-for lines and savings is cross-fitted: the predictor comes from no-evidence runs 1--2 and the outcome from runs 3--4 against all runs with the package, and vice versa; the partial correlation additionally controls for the predictor half's read and search count. A manual audit of 12 runs with the package, sampled with a fixed seed, classified where the edit site came from: a package candidate (5), the issue text itself (3) or neither (4).

\section{Distilling Efficient Behavior}
\label{app:distill}

If evidence mainly teaches the agent to stop searching sooner, could a model learn that behavior directly? We fine-tune Qwen3.5-9B on trajectories of Qwen3.6-27B, the setting studied by \citet{chen2025locagent} for localization. Teacher runs come from the 224 Loc-Bench tasks, disjoint from evaluation. A run is selected, by rules fixed before sampling, if it submits a non-empty patch that edits a gold-edited function, has no turn without a command, and costs no more than the median of its task and condition; we keep at most two per task and condition and balance the two conditions, giving 464 trajectories. Runs whose package was replaced are split so that every reply is trained in the context it was generated in. We train a LoRA adapter (rank 16, two epochs) once, without tuning, and evaluate base and fine-tuned models on the 80 Verified tasks with and without the package (\cref{app:training}).

\begin{figure*}[h]
\centering
\figorbox{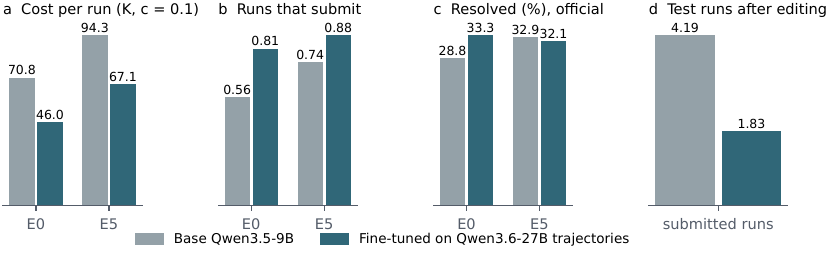}{\textwidth}
\caption{Base Qwen3.5-9B versus the same model fine-tuned on efficient Qwen3.6-27B trajectories, on SWE-bench Verified. (a) Cost per run at $c=0.1$; (b) share of runs that submit; (c) official resolved rate; (d) script or test runs after the first edit, among submitted runs.}
\label{fig:distill}
\end{figure*}

\paragraph{Efficiency transfers.}
Fine-tuning cuts cost by 32\% ($-26.0$K tokens per run, averaged over both conditions), shortens runs by 9.7 turns, and raises the share of runs that submit by 20 points (\cref{fig:distill}).

\paragraph{Quality does not.}
The official resolved rate is unchanged: $+4.5$ points without the package (16 tasks better, 9 worse) and $-0.6$ with it. Among submitted runs it falls from 0.43--0.44 to 0.37--0.38. The student submits more because it checks less: after its first edit it runs a script or test 1.83 times instead of 4.19, and 34\% of its submissions follow an edit with no run at all, against 26\% for the base model. Our three preregistered hypotheses, that fine-tuning would change how much the student benefits from the package, were all unsupported.

The two results mirror each other. The package and the fine-tuned model both make the agent stop exploring sooner, and in both cases the patches are no better: what the agent saves is search effort, not the reasoning that decides whether an edit is correct.

\section{Fine-Tuning Details}
\label{app:training}
Teacher runs are Qwen3.6-27B on 224 Loc-Bench tasks, four runs per task with and without the package, sampled with full transcripts. Selection rules (\cref{app:distill}) yield 232 runs per condition and 696 training samples after splitting runs at the package replacement. We train a LoRA adapter (rank 16, $\alpha=32$, attention and MLP projections) for two epochs with learning rate $10^{-4}$, linear warm-up over ten updates and cosine decay to $10^{-5}$, effective batch size 8, on sequences up to 32{,}768 tokens; the loss is computed on assistant replies only. Base and fine-tuned models are served together and evaluated in the same batch with the same seeds, two runs per task and condition. The preregistered interaction tests were: fine-tuning reduces exploration more with the package than without ($+2.13$); it raises on-target accuracy more with the package ($-0.062$); it reduces cost more with the package ($-2{,}299$ tokens, $[-15{,}435, +10{,}878]$). None was supported.

\section{Measurement Protocol}
\label{app:protocol}
\Cref{tab:protocol} collects the measurement biases we found, with their size in our runs. A cost comparison of agent methods should (i) report uncached, cached and output tokens separately; (ii) count each run's own cache hits at token granularity; (iii) state the server's cache block size and whether it reuses a previous reply; (iv) not run conditions that share prefixes side by side, or correct for it; (v) report evaluation noise from repeated runs, or compare on the same recorded runs; and (vi) state conclusions for a price region, with joint inference over it (\cref{app:pricemap}).
\begin{table}[h]
\centering
\caption{Measurement biases in cost experiments, with their size in our runs.}
\label{tab:protocol}
\footnotesize
\resizebox{\columnwidth}{!}{%
\begin{tabular}{@{}p{2.6cm}p{5.4cm}@{}}
\toprule
Bias & Size in our runs \\
\midrule
Concurrent conditions share caches & Verified package premium $+8\%$ concurrent, $+29\%$ isolated \\
Block-granular hit counts & 9.8\% fewer hits than the token count (784-token blocks); rolling masking $+173\%$ instead of $+116\%$ at $c=0.1$ \\
Reuse of the previous reply & never for Qwen3.6-27B (0 of 27{,}389 isolated requests, history re-rendered); 83\% of requests for Qwen3-Coder; counting it moves rolling masking from $+87\%$ to $+98\%$ \\
Evaluation noise, cost & relative standard error of one evaluation's mean cost 3--4\% (201 tasks, one run each); two evaluations differ by 4--5\% \\
Evaluation noise, resolution & two evaluations of one agent differ by 1.6 points (178 tasks, two runs each) \\
\bottomrule
\end{tabular}}
\end{table}

\section{Price Maps}
\label{app:pricemap}
For the 201-task re-evaluation we bootstrapped (2{,}000 resamples over tasks) each method's relative cost difference to no management on the grid $c\in\{0.025, 0.05, 0.1, 0.25, 0.5, 1\}$, $p\in\{2,4,6,8\}$, with simultaneous 95\% intervals from the maximum $|t|$ over the grid (critical values 2.07--2.31 instead of 1.96). On the current range, a one-sided bound on the maximum over its four corners tests "cheaper throughout" and a bound on the minimum tests "more expensive throughout". No method was cheaper throughout for any of the three models. Rolling masking was more expensive throughout for all three, summarization and AgentDiet (extra calls at full price) for Qwen3.6-27B and Qwen3-Coder, and AgentDiet with extra calls at a fifth of the price for Qwen3.6-27B; batched and break-even masking were inconclusive. Outside the range, batched and break-even masking were significantly cheaper for Qwen3.6-27B from $c=0.25$ at every $p$.

\section{What the Agent Does After Masking}
\label{app:reaction}
Paired by task and seed with the unmasked run, rolling, batched and break-even masking added 5.3, 2.9 and 2.6 commands per run for Qwen3.6-27B: $+3.0$, $+1.7$ and $+1.5$ reads, $+2.1$, $+1.0$ and $+1.5$ searches, and no more edits. Unmasked runs already re-read a file they had read 8.1 times per run; masking added 3.1, 1.8 and 1.9 such re-reads. Priced at each masked run's mean cost per command, the extra commands amount to $+23.5\%$, $+6.2\%$ and $+6.2\%$ of the unmasked cost at $c=0.1$, against a behavioral term of $+25.7\%$, $+8.2\%$ and $+7.1\%$ in the fixed-action replay. For Qwen3-Coder the change was $-0.5$ to $+1.8$ commands. We also replayed the unmasked runs with a rule that masks only provably stale observations (output of a read whose file was later edited, or of a command later re-run), ten at a time, keeping the latest ten (preregistered). Such observations held 12\% of the observation tokens for Qwen3.6-27B; the rule masked 19\% of what batched masking does and changed cost by $+1.4\%$, $-1.9\%$ and $-4.6\%$ at $c=0.1$, 0.25 and 1 with actions fixed, against $-3.4\%$, $-17.4\%$ and $-29.0\%$ for batched masking (Qwen3-Coder: 17\%, $+1.3\%$ to $-2.7\%$). It fell short of the preregistered 25\% needed for an online pilot.

\section{Coverage of Additional Official Evaluations}
\label{app:officialreturn}
The additional Windows SWE-bench harness package contains context-management runs as well as stopping runs. It returned 1,338 of 1,341 distinct non-empty patch reports across 451 tasks. Gold self-checks pass on 430 tasks, fail on seven and are missing on fourteen. The three missing model-patch reports have recorded application, timeout or image-download failures; their outcomes remain unknown rather than being counted as model failures. Empty model patches remain unresolved under the existing protocol.

We matched local and official results by task ID and patch hash, restricting comparison to passing official gold checks. Of 564 patches with unambiguous local scores, 554 agree. Another 691 eligible patches lack an unambiguous local score and are excluded from this agreement calculation. This checks the returned patches, not whether patch extraction captured every edit. The table below reports only context interventions relevant to this paper.

\paragraph{Paired official comparisons.}
We retain only tasks with passing gold checks and all required reports in both conditions, average repeats within tasks, and use 2,000 paired task bootstrap samples (seed 20261001). These are supplementary official analyses on their stated task subsets. They do not replace local results on a different denominator. Missing reports remain unknown.
\begin{center}
\small
\setlength{\tabcolsep}{2pt}
\begin{tabular}{lrr}
\toprule
Comparison & Tasks & Change (pp) \\
\midrule
MAR025, Qwen3.6 & 182 & $+2.2$ \\
MAR025, Coder & 182 & $+1.1$ \\
Isolated E5 vs. E0 & 77 & $+1.3$ \\
MDc vs. MR & 183 & $+0.5$ \\
\bottomrule
\end{tabular}
\end{center}

The paired bootstrap tests do not establish a positive effect. These official results do not establish a resolution gain from the corresponding context interventions. Cost estimates depend on the recorded token streams and are unchanged by official patch scoring.